\documentclass[twoside]{article}

\usepackage[accepted]{aistats2023}
\usepackage{amsmath,amsfonts,bm}

\def\eqref#1{equation~\ref{#1}}

\def\1{\bm{1}}

\DeclareMathAlphabet{\mathsfit}{\encodingdefault}{\sfdefault}{m}{sl}
\SetMathAlphabet{\mathsfit}{bold}{\encodingdefault}{\sfdefault}{bx}{n}

\newcommand{\sigmoid}{\sigma}

\usepackage{hyperref}
\usepackage{url}

\usepackage{tikz}
\usepackage{amsmath}

\usepackage{microtype}
\usepackage{booktabs} 

\usepackage{amsmath}
\usepackage{amssymb}
\usepackage{mathtools}
\usepackage{amsthm}

\usepackage[capitalize,noabbrev]{cleveref}

\usepackage[textsize=tiny]{todonotes}

\newcommand{\calL}{\mathcal{L}}
\newcommand{\calM}{\mathcal{M}}
\newcommand{\calT}{\mathcal{T}}

\usepackage[inline]{enumitem}
\usepackage{pifont}
\usepackage{wrapfig}
\usepackage[utf8]{inputenc}
\usepackage{amsmath}
\usepackage{amsfonts}
\usepackage{booktabs} 
\usepackage{xcolor}
\usepackage{comment}
\usepackage{multirow}
\usepackage{amssymb}
\usepackage{mathtools,amsthm}
\usepackage[noend]{algorithm}
\usepackage[algo2e]{algorithm2e}
\usepackage[noend]{algorithmic}
\usepackage{epstopdf}
\usepackage{amsthm}
\usepackage{enumitem}
\usepackage{lipsum}
\usepackage{soul}
\usepackage{subcaption}
\usepackage{framed}
\usepackage{times}
\usepackage{diagbox}
\usepackage{empheq}
\usepackage{cases}

\usepackage[export]{adjustbox}
\usepackage{bm}

\usepackage{soul}
\usepackage{comment}
\usepackage{caption}
\newtheorem{mydef}{Definition}
\newtheorem{theorem}{Theorem}

\newtheorem{corollary}{Corollary}

\makeatletter
\def\ps@justfooters{\let\@mkboth\@gobbletwo\def\@oddhead{}%
	\def\evenhead{}}
\makeatother

\newcommand{\procfont}[1]{\textbf{#1}}

\newcommand{\bits}{{\{0,1\}}}

\newlength{\saveparindent}
\newlength{\saveparskip}
\newcounter{ctr}

\newcounter{ectr}

\newlength{\savejot}
\newcommand{\calD}{{\cal D}}

\newcommand{\calN}{{\cal N}}

\newcommand{\getsr}{{\:\stackrel{{\scriptscriptstyle \hspace{0.2em}\$}} {\leftarrow}\:}}

\makeatletter

\newcommand{\adversary}[1]{\underline{\procfont{adversary} {#1}:}}

\newcommand{\experimentv}[1]{\underline{{#1}}\smallskip}

\newcommand{\truc}[1]{{\color{magenta} #1}}
\newcommand{\cmt}[1]{{\color{blue} \footnotesize \# #1}}

\newcommand{\calA}{\mathcal{A}}

\newcommand{\calY}{\mathcal{Y}}

\DeclareMathOperator{\relu}{ReLU}
\DeclareMathOperator{\Range}{Range}

\begin{document}

\twocolumn[

\aistatstitle{Active Membership Inference Attack under Local Differential Privacy in Federated Learning}



\aistatsauthor{ Truc Nguyen \And Phung Lai \And  Khang Tran}

\aistatsaddress{University of Florida\\truc.nguyen@ufl.edu \And New Jersey Institute of Technology\\tl353@njit.edu \And New Jersey Institute of Technology\\kt36@njit.edu} 

\aistatsauthor{ NhatHai Phan \And My T. Thai\textsuperscript{*}}

\aistatsaddress{New Jersey Institute of Technology\\phan@njit.edu \And University of Florida\\mythai@cise.ufl.edu}

]

\runningauthor{Truc Nguyen, Phung Lai, Khang Tran, NhatHai Phan, My T. Thai}

\begin{abstract}

    Federated learning (FL) was originally regarded as a framework for collaborative learning among clients with data privacy protection through a coordinating server. In this paper, we propose a new active membership inference (AMI) attack carried out by a dishonest server in FL. In AMI attacks, the server crafts and embeds malicious parameters into global models to effectively infer whether a target data sample is included in a client's private training data or not. By exploiting the correlation among data features through a non-linear decision boundary, AMI attacks with a certified guarantee of success can achieve severely high success rates under rigorous local differential privacy (LDP) protection; thereby exposing clients' training data to significant privacy risk. Theoretical and experimental results on several benchmark datasets show that adding sufficient privacy-preserving noise to prevent our attack would significantly damage FL's model utility.
\end{abstract}

\section{INTRODUCTION}
Federated Learning (FL) has emerged as a promising large-scale collaborative learning framework in recent years. 
By design, FL enables participating clients to collaboratively train a global model through a coordinating server. Although training data never leaves clients' devices, a dishonest server can still infer the membership information of any client's training data through observing their local model updates by using (passive or active) membership inference attacks \cite{shokri2017membership,salem2018ml,song2021systematic,DBLP:conf/sp/NasrSH19}. 
For that reason, FL in its primitive form offers little to no privacy protection. 

To address the problem, several privacy-preserving mechanisms, such as local differential privacy (LDP), have been developed to challenge membership inference (MI) attacks in general and active membership inference (AMI) attacks in particular by effectively protecting the membership information of client's training data with upper-bounded privacy leakage \cite{arachchige2019local,sun2020ldp,lai2021bit,lyu2020towards}. Recent studies apparently show that LDP protection is effective in mitigating MI and AMI attacks \cite{rahman2018membership, 10.1007/978-3-030-81242-3_2, GU2022103201}.
The key reason for this result is that existing attacks have not fully conveyed privacy risks in FL by under-exploiting the correlation among data features and LDP protection. That poses previously 
unexplored privacy risk to the clients' local training data.




\textbf{Key Contributions.} To tackle that problem, we 
first formalize a new AMI threat model equipped with an AMI attack from a dishonest server. The key idea is that, given a target data sample, the server carefully crafts malicious weights of the global model such that the model updates from the clients would expose the membership information of the target data sample through the behavior of a chosen neuron. A chosen neuron is only activated given the target data sample controlled by a non-linear decision boundary embedded inside the malicious weights. With 
our non-linear decision boundary, the server can infer this membership information with severely high success rates. Furthermore, the server effortlessly achieves this result with a minimal change to the global model parameters within one training iteration. 


In addition, we take a step forward and devise an AMI attack strategy under LDP protection to significantly amplify the privacy risk in FL. By adding a certain amount of privacy-preserving noise to the local data before training, LDP can protect the data
with formal privacy leakage bounds (controlled by a privacy budget $\varepsilon$) \cite{dwork2014algorithmic}. The key advantage of our AMI attack is exploiting the correlation among data features to distinguish the target data sample from others under LDP protection (Eq. \ref{equ:cond-ldp}). If LDP-preserving noise is insufficient to break this correlation (i.e., large privacy budgets $\varepsilon$), clients' local data will be at risk of our AMI attack with certified guarantees of success. Meanwhile, large privacy-preserving noise (i.e., small privacy budgets $\varepsilon$) can significantly damage the FL's model utility.

Our theoretical and experimental results in many benchmark datasets show that our AMI attacks stress-test the fundamental trade-offs between model utility and privacy risk in FL to a new level. This is reflected through notably high success rates under rigorous LDP protection (i.e., small privacy budgets $\varepsilon$ which significantly degenerate FL's model utility).



\paragraph{Organization.} The remainder of the paper is structured as follows. Section \ref{sec:prelim} presents background and establishes our threat model. Section \ref{sec:amia} introduces our AMI attack from a dishonest server in FL. We devise an attack strategy under LDP protection with certified guarantees of success in Section \ref{sec:ldp}. Section \ref{sec:eval} evaluates the performance of AMI attacks in several benchmark datasets. Section \ref{sec:related} discusses related work and Section \ref{sec:conclude} provides concluding remarks.

\section{BACKGROUND AND THREAT MODEL}\label{sec:prelim}

In this section, we briefly review the background of federated learning and differential privacy, and then introduce our active membership inference threat model.

\subsection{Background}

\paragraph{Federated Learning (FL).}
We focus on a horizontal setting of FL in which different clients hold the same set of features but different sets of samples. 
We denote $f_\theta : \mathbb{R}^d \rightarrow \mathbb{R}^k$ as a $k$-class neural network model that is parameterized by a set of weights $\theta$. The aim of $f_\theta$ is to map a data point $x \in \mathbb{R}^d$ to a vector of posterior probabilities $f_\theta(x) = \calY$ over $k$ classes, where the sum of all values in $\calY$ is 1.

FL 
is an iterative learning framework for training a global model $f_\theta$ on distributed data owned by $N$ different clients $\{u_j\}_{j=1}^N$.  A central server coordinates the training of $f_\theta$ by iteratively aggregating gradients
computed locally by the clients. Let $i \in \mathbb{Z}_{\geq 0}$ be the current iteration of the FL protocol, and $\theta^i$ be the set of parameters at iteration $i$. At iteration $i=0$, the global $\theta^i$ is initialized randomly by the central coordinating server. At every iteration $i$, a subset of $M < N$ clients is 
randomly selected to participate in the training. Each of the selected clients $u_j$ receives $f_{\theta^i}$ from the central server and calculates the gradients $G_j^i$ for $f_{\theta^i}$ using their local training batch $\calD_j$. Specifically, $G_j^i = \nabla_{\theta^i} \calL(\calD_j, \theta^i)$ where $\calL$ is a loss function. Then, each $u_j$ uploads its gradients to the central server, who averages all of these gradients to compute the global model’s parameters with a learning rate $\eta$:

\begin{equation}
    G^{i} = \frac{1}{M}\sum_{j=1}^M G_j^{i}, \quad \theta^{i+1} = \theta^i - \eta G^i
\end{equation}
 The training continues until $f_{\theta^i}$ converges.

\paragraph{FL with Local Differential Privacy (LDP).}
Recent attacks have shown that clients' training data samples can be extracted from the shared gradients \cite{NEURIPS2019_60a6c400,yin2021see}. These attacks underscore privacy risks in FL. Therefore, privacy-preserving mechanisms are needed to control and mitigate the privacy risks introduced by gradient sharing while optimizing utility. 

Local differential privacy (LDP) \cite{dwork2006calibrating, erlingsson2014rappor} is one of the auspicious solutions, given its formal  protection without an undue sacrifice in computation efficiency. LDP builds on the ideas of \textit{randomized response} \cite{warner1965randomized}, which was initially introduced to allow survey respondents to provide their  inputs while maintaining their confidentiality. 
The definition of $\epsilon$-LDP is as follows:

\begin{mydef}{$\epsilon$-LDP.} A randomized algorithm $\mathcal{M}$ fulfills $\epsilon$-LDP, if for any two inputs $x$ and $x'$, and for all possible outputs $\mathcal{O} \in \Range(\mathcal{M})$, we have:
$Pr[\mathcal{M}(x) = \mathcal{O}] \leq e^{\epsilon} Pr[\mathcal{M}(x') = \mathcal{O}]$,
where $\epsilon$ is a privacy budget and $\Range(\mathcal{M})$ denotes every possible output of  $\mathcal{M}$.
\label{Different Privacy} 
\end{mydef}

The privacy budget $\epsilon$ controls the amount by which the distributions induced by inputs $x$ and $x'$ may differ. A smaller value of $\epsilon$ enforces a stronger privacy guarantee but reduces model utility. 

\subsection{Active Membership Inference Threat Model}\label{ssec:threatmodel}
Previous studies typically focus on a scenario in which the central server is interested in uncovering client information by examining local updates from the clients, but still abiding by the system protocol. This threat model is commonly referred to as honest-but-curious or semi-honest. However, 
this threat model 
undermines the vulnerability of the FL system as in practice, the server can 
deviate from the protocol to strengthen the privacy attacks \cite{boenisch2021curious,nguyen2022blockchain,fowl2021robbing}. In this work, we are thus interested in explicitly malicious (or actively dishonest) servers that may modify the model architecture and/or model parameters before dispatching them to the clients. In this regard, we propose an active membership inference threat model, in which a dishonest server maliciously adjusts the model parameters to determine whether a target data sample is in the local training dataset of a client.

\begin{figure}[t]
\begin{framed}
\begin{flushleft}
\experimentv{$\mathsf{Exp}{(\mathcal{A},\mathcal{L},\mathbb{D})}$:}\\
$\mathcal{D} \sim \mathbb{D}^n$ \cmt{Sample $n$ data points from $\mathbb{D}$ into $\calD$}\\
$b \getsr \bits$ \cmt{Flip a bit $b$ uniformly at random}\\
\If{$b=1$} {
$t \getsr \mathcal{D}$ \cmt{Choose $t$ uniformly from $\calD$}
}
\Else{$t \sim \mathbb{D} \setminus \mathcal{D}$ \cmt{Sample $t$ from $\mathbb{D}$ s.t. $t \notin \mathcal{D}$}} 

$\theta \gets \mathcal{A}^\mathbb{D}_{\mathsf{INIT}}(t)$ \cmt{The adversary receives $t$ and returns a set of parameters $\theta$}\\
$G \gets \nabla_\theta \calL(\calD, \theta)$ \cmt{Compute the gradients from $\theta$ and $\calD$}\\
$b' \gets \mathcal{A}^\mathbb{D}(t, G)$ \cmt{The adversary receives $t, G$ and returns a bit $b'$}\\
\textbf{Ret} $[b'=b]$ \cmt{The game returns 1 if $b'=b$ (the adversary wins), 0 otherwise}
\end{flushleft}
\end{framed}
\caption{AMI Threat Model as a Security Game.}
\label{fig:threat-model}
\end{figure}

We describe the active membership inference threat model as follows. We denote $\calA$ as the central server in FL, which is also the adversary. Note that this threat model represents an attack at an arbitrary iteration that targets a specific client. Let $\mathcal{D} = \{(x_i,y_i)\}$ be the batch of training data of the target client. The set $\calD$ contains sample $x_i\in \mathbb{R}^d$ and its ground-truth label $y_i \in \{1, ..., k\}$ with $k$ classes. Suppose that $\calD$ is sampled from a distribution $\mathbb{D}$ on $(x_i, y_i)$ that the adversary $\calA$ has knowledge of (i.e., similar to existing studies \cite{carlini2022membership,yeom2018privacy,shokri2017membership}).
This is practical in the real world since the server can collect a massive amount of data that covers the local data distribution of a sufficient number of clients \cite{shokri2017membership}. 
The adversary outputs maliciously crafted model parameters $\theta$ to the target client. The client sends the local gradients $G = \nabla_\theta \calL(\calD, \theta)$ back to the adversary. By observing the local gradients $G$, the goal of the server's attack is to determine whether a target (data) sample $t \in \mathbb{R}^d$ is included in the local training set $\calD$.
More formally, the adversary can be defined as the following function:
\begin{equation}\label{equ:adv}
    \calA^\mathbb{D}: t, G \rightarrow \{0,1\}
\end{equation}
where $\calA^\mathbb{D}$ denotes the query access to $\mathbb{D}$, 1 means $t\in \calD$, and 0 otherwise.


We formalize this threat model as a security game $\mathsf{Exp}{(\mathcal{A},\mathcal{L},\mathbb{D})}$ between a challenger and the adversary in Fig. \ref{fig:threat-model}. From that, the adversary's advantage, or the attack success rate, is defined as follows:
\begin{equation}\label{equ:suc}
\begin{aligned}
      \text{Adv}^\mathcal{A} &= \Pr[\mathsf{Exp}{(\mathcal{A},\mathcal{L},\mathbb{D})} = 1]\\
      &= \frac{1}{2}\Pr[b'=1|b=1] + \frac{1}{2}\Pr[b'=0|b=0]
\end{aligned}
\end{equation}
where $\Pr[b'=1|b=1]$ is the True Positive Rate (TPR), and $\Pr[b'=0|b=0]$ is the True Negative Rate (TNR). The success rate $\text{Adv}^\mathcal{A}$ should be greater than $0.5$, which is the probability of random guessing.



\section{ACTIVE MEMBERSHIP INFERENCE (AMI) ATTACK}\label{sec:amia}

This section first discusses the technical intuition of membership inference through gradients. Based on this concept, we then describe our proposed strategy to launch the AMI attack from a dishonest server.

\subsection{Inferring Membership via Gradients}
As shown 
in Fig. \ref{fig:threat-model}, the adversary $\calA$ receives the gradients $G$ that was computed on the training set $\calD$, and 
$\calA$ wishes to determine whether $t\in \calD$. This section discusses how the membership information can be inferred through gradients.

Suppose that, on an input data point $x \in \mathbb{R}^d$, the output of the \textit{\textbf{first fully-connected layer}} is expressed as $\relu(Wx + b) = \max(0, Wx + b)$ where $W\in \mathbb{R}^{r\times d}$ is the weight matrix of that layer and $b\in \mathbb{R}^r$ is the bias vector ($r$ is the number of neurons in the layer). To express the output of the $i$-th neuron of that layer, we denote $W_i$ as the corresponding row in the weight matrix and $b_i$ as the corresponding component in the bias vector. We observe that, when $W_i x + b_i \leq 0$, the ReLU outputs zero, in other words, the neuron $i$ is not activated by $x$. As a result, the gradient of neuron $i$, denoted by $G^{(x)}_i$, is zero at the data point $x$. Otherwise, when $W_i x + b_i > 0$, the gradient $G^{(x)}_i$ is non-zero. 

As the gradient $G$ is computed over the whole training set $\calD$, the gradient of a neuron $i$ received by the adversary is the average of gradients over all data points $x\in \calD$, i.e., $G_i = \frac{1}{|\calD|} \sum_{x\in \calD} G^{(x)}_i $. If there exists a neuron $i$ that is \textit{\textbf{activated only}} by a target data sample $t$ ($t \in \mathbb{R}^d$), and \textit{\textbf{not activated}} by any other data samples $x \neq t$, then we have:
\begin{equation}\label{equ:linear}
\left \{
  \begin{aligned}
    &\sum_{j = 1}^d W_{ij} t_j > 0 \\
    &\sum_{j = 1}^d W_{ij} x_j \leq 0, \quad \forall x\in \calD \setminus t
  \end{aligned} \right.
\end{equation}
note that we suppress the bias term for simplicity. 

If $t\in \calD$, then $G_i$ will be non-zero; otherwise, $t\notin \calD$ results in $G_i$ being zero. From that, the adversary upon seeing $G_i$ can easily infer whether the target data sample $t$ was a part of the training set $\calD$ or not. The formulation in Eq. (\ref{equ:linear}) is similar to the framework proposed by \cite{boenisch2021curious} for conducting data reconstruction attacks where a neuron that is activated only by one sample can be used to perfectly reconstruct that sample. However, we shall see below that this simple formulation is actually inapplicable to the membership inference attack, thus requiring a more advanced strategy to launch the attack.

\subsection{Attack Strategy: Manipulating Model Parameters via Training a Chosen Neuron}\label{ssec:strat}


As aforementioned, if there exists a neuron 
that is activated only by the target data sample $t$, it is sufficient to determine whether $t$ is in the training set $\calD$ or not.
From the threat model, the adversary can determine the model parameters $\theta$, which includes the weight matrix $W$. Obviously, the adversary can choose some neuron $i$ and try to solve Eq. (\ref{equ:linear}) for $W$ to realize the conditions of the chosen neuron. Since the adversary does not know $\calD$, it can only approximate Eq. (\ref{equ:linear}) for all of its observed data samples $x$ that are different from $t$ (i.e., $x \neq t$ and $x \notin \mathcal{D}$). However, that makes Eq. (\ref{equ:linear}) infeasible due to the linearity of the functions $\sum_{j = 1}^d W_{ij} t_j$ and $\sum_{j = 1}^d W_{ij} x_j$, as shown in Appendix \ref{app:linear}.







To address this issue, we introduce non-linearity into the equation. To do so, instead of relying on the first layer, the adversary can choose a neuron in the \textit{\textbf{second fully-connected layer}} such that the neuron is activated only by the target data sample $t$, and not activated by any other data samples $x \neq t$. Let us denote $h \in \mathbb{R}^r$ as the weight vector of the chosen neuron in the second layer, the attack is successful if we can find $(h, W)$ such that:
\begin{equation}\label{equ:cond-nonlinear}
\left \{
  \begin{aligned}
    &\sum_{i=1}^r h_i \relu(\sum_{j=1}^d W_{ij}t_j) > 0 \\
    &\sum_{i=1}^r h_i \relu(\sum_{j=1}^d W_{ij}x_j) \leq 0, \quad \forall x \neq t
  \end{aligned} \right.
\end{equation}

To solve Eq. (\ref{equ:cond-nonlinear}), we can train the chosen neuron to be activated only by the target data sample $t$. For the training, we first put forth a logistic sigmoid function ($\sigmoid$) on the output of the chosen neuron. As a result, the function of the chosen neuron becomes: 
\begin{equation}\label{equ:target}
\begin{aligned}
    s(x) &= \sigmoid(\sum_{i=1}^r h_i \relu(\sum_{j=1}^d W_{ij}x_j)) \\
    &= \sigmoid(h \cdot \relu(Wx))
\end{aligned}
\end{equation}

Next, we sample a dataset $\mathcal{X} \sim \mathbb{D}^m$, in which  we assign a label $1$ for the target data sample $t$ and label $0$ for all other data samples $x\in \mathcal{X}\setminus t$. After that, we train the chosen neuron using cross-entropy loss. The key idea is that the training process tries to make $s(t) = 1$, and $s(x) = 0$ for $x\neq t$. When $s(t) > 0.5$, it means that $h \cdot \relu(Wt) > 0$; otherwise, when $s(x) < 0.5$, we have $h \cdot \relu(Wx) < 0$ and that conforms to Eq. (\ref{equ:cond-nonlinear}). 

When the adversary receives the gradient $G$, it can observe the gradient of that chosen neuron to determine whether the target data sample $t$ is in the training set $\calD$ ($t \in \calD$) or not, as discussed in the previous section. In particular, the adversary extracts the gradient of the chosen neuron, denoted by $g_t$, from $G$ and sees whether $g_t$ is non-zero. \textit{\textbf{If $g_t$ is zero}}, the adversary predicts that \textit{\textbf{the target data sample $t$ is not in the training set $\calD$}} (i.e., $t\notin \calD$). This is because the chosen neuron was not activated during the gradient computation on the training set $\calD$. Otherwise, \textit{\textbf{if $g_t$ is non-zero}}, then the adversary predicts that \textit{\textbf{the target data sample $t$ is in the training set $\calD$}} (i.e., $t\in \calD$). 

Fig. \ref{fig:strategy} shows a design of the adversary $\calA$ according to the threat model in Fig. \ref{fig:threat-model}. Note that our attack strategy only modifies the parameters of 1 chosen neuron in the second layer and $r$ associated neurons in the first layer. That makes our attack feasible by enabling us to make a minimal change to the model parameters, and the attack can be carried out within one FL training iteration.

\begin{figure}[t]
\small
\center
\begin{framed}
  {
    \begin{flushleft}
    \adversary{$\calA^{\mathbb{D}}_{\mathsf{INIT}}(t)$}\\
    $\mathcal{X} \sim \mathbb{D}^m$\\
    $D_{A} \gets \bigcup_{x \in \mathcal{X} \setminus t} \{(x, 0)\} $\\
    $D_{A} \gets D_{A} \cup \{(t, 1)\}$\\
    Train $h, W$ (Eq. \ref{equ:target}) from dataset $D_{A}$\\
    Initialize $\theta$\\
    $\theta \gets \theta \cup (h, W)$\\
    \textbf{Ret} $\theta$
    \end{flushleft}
  }
  {
  \begin{flushleft}
    \adversary{$\calA^{\mathbb{D}}(t, G)$}\\
    Extract $g_t$ as the gradient of the chosen neuron from $G$\\
    \textbf{Ret} $[g_t \neq 0]$
    \end{flushleft}
  }
  \label{fig:sfig2}
 \end{framed}
 \caption{AMI Attack Strategy of the Adversary $\calA$.}
\label{fig:strategy}
\end{figure}







\section{AMI ATTACK UNDER LDP WITH CERTIFIED GUARANTEE OF SUCCESS}\label{sec:ldp}
LDP is generally regarded as an effective defense against privacy inference attacks given its rigorous privacy protection compared with other approaches \cite{wagh2021dp}. To tolerate such mechanisms,  our AMI attack exploits the correlation among input features captured through a non-linear decision boundary 
to distinguish the target data sample $t$ from others (Eq. \ref{equ:cond-nonlinear}) under LDP protection. If LDP-preserving noise is insufficient to break this correlation (i.e., large privacy budgets $\varepsilon$), clients' local data will be at risk of our AMI attack. Meanwhile, large privacy-preserving noise (i.e., small privacy budgets $\varepsilon$) can significantly damage the model utility. In FL, it is challenging for clients to identify suitable privacy budgets given their limited local training data. Therefore, they usually rely on the server to provide the privacy budget $\varepsilon$ and the LDP-preserving mechanism $\calM$. That increases the risk of exposing their local training data to a dishonest server under our AMI attack.
We focus on shedding light on the fundamental trade-offs between $\varepsilon$-LDP protection, model utility, and privacy risk with a certified bound for our AMI attack to be successful.



Each client independently perturbs every training data sample in their local training data $\calD$ using an LDP-preserving mechanism $\calM$ (Def. \ref{Different Privacy}) to obtain a randomized local training set $\calD' = \calM(\calD, \varepsilon) = \{\calM(x, \varepsilon)\}_{x\in \calD}$. The client sends the gradients derived from the randomized training set $G=\nabla_\theta \calL(\calD', \theta)$ to the server. The mechanism $\calM$ and the privacy budget $\varepsilon$ are known to the adversary $\calA$, thus, the adversary function in Eq. (\ref{equ:adv}) is re-defined under LDP context as $\calA^{\mathbb{D},\calM}_{LDP}: t, G, \varepsilon \rightarrow \{0,1\}$. We enhance the threat model to reflect the use of LDP in Fig. \ref{fig:threat-model-ldp}.


\paragraph{Attack Strategy.} Given the gradients $G$ computed on LDP-perturbed data $\calD'$, applying the same attack strategy that was discussed in Section \ref{ssec:strat} would not work effectively. Suppose that the target sample $t$ is in $\calD$, and that the adversary were able to train a neuron that is activated only by the target sample $t$, i.e., $h\cdot \relu(Wt) > 0$ and $h\cdot \relu(Wx) < 0$ for $x\neq t$. However, it is very likely that $t$ is not in the randomized local training set $\calD'$ since $t$ was randomized under LDP. As the client uses $\calD'$ for gradients computation, the chosen neuron may remain inactivated, i.e.,  $h\cdot \relu(W\cdot \calM(t, \varepsilon)) < 0$. Hence, the attack fails to infer the correct membership of $t$.

Therefore, it is necessary for the adversary to ensure that the chosen neuron is activated only by the randomized target sample $\calM(t, \varepsilon)$. Similar to Eq. (\ref{equ:cond-nonlinear}), we formulate this observation as finding $(h,W)$ such that:
\begin{equation}\label{equ:cond-ldp}
\left \{
  \begin{aligned}
    &h \cdot \relu(W\calM(t, \varepsilon)) > 0 \\
    &h \cdot \relu(Wx) \leq 0, \quad \forall x \neq \calM(t, \varepsilon)
  \end{aligned} \right.
\end{equation}




To develop an effective attack for Eq. (\ref{equ:cond-ldp}), it is essential for the chosen neuron to be activated if the client uses $\calM(t, \varepsilon)$ regardless of the randomness introduced by the LDP-preserving mechanism $\calM(\cdot, \varepsilon)$. To achieve our goal, we strengthen our attack by generating a set $\calT$ of $l$ perturbations $\calM(t, \varepsilon)$, that is, invoking $\calM(t, \varepsilon)$ $l$ times with independent draws of LDP-preserving noise. Next, we sample a training set $\mathcal{X} \sim \mathbb{D}^m$ such that $ \mathcal{X} \cap \calT = \emptyset$. Then, we assign label $1$ for 
samples in $\calT$ (which contains randomized versions of  $t$) and label $0$ for samples in $\mathcal{X}$. Finally, we train the chosen neuron using cross-entropy loss. The remaining steps follow Section \ref{ssec:strat}. Fig. \ref{fig:strategy-ldp} shows the design of the adversary $\calA_{LDP}$ with respect to the threat model in Fig. \ref{fig:threat-model-ldp}.


\begin{figure}[t]
\small
\begin{framed}
\begin{flushleft}
\experimentv{$\mathsf{Exp}_{LDP}{(\mathcal{A}_{LDP},\mathcal{L},\mathbb{D}, \calM, \varepsilon)}$:}\\
$\mathcal{D} \sim \mathbb{D}^n$ \\
$b \getsr \bits$ \\
\If{$b=1$} {
$t \getsr \mathcal{D}$
}
\Else{$t \sim \mathbb{D} \setminus \mathcal{D}$ } 
$\theta \gets \mathcal{A}^{\mathbb{D},\calM}_{LDP, \mathsf{INIT}}(t,\varepsilon)$ \\
$\calD' \gets \calM(\calD, \varepsilon)$ \cmt{Apply the LDP mechanism on $\calD$}\\
$G \gets \nabla_\theta \calL(\calD', \theta)$ \\
$b' \gets \mathcal{A}^{\mathbb{D}, \calM}_{LDP}(t, G, \varepsilon)$ \\
\textbf{Ret} $[b'=b]$ 
\end{flushleft}
\end{framed}
\caption{AMI Threat Model under LDP Mechanisms.}
\label{fig:threat-model-ldp}
\end{figure}

\begin{figure}[h]
\small
\center
\begin{framed}
  {
    \begin{flushleft}
    \adversary{$\calA^{\mathbb{D},\calM}_{LDP,\mathsf{INIT}}(t,\varepsilon)$}\\
    Choose $l\in \mathbb{N}$\\
    $\calT \gets \emptyset$\\
    \For{$i=1$ to $l$}
    {$\calT\gets \calT \cup \{\calM(t,\varepsilon)\}$ }

    $\mathcal{X} \sim \mathbb{D}^m \setminus \calT $ \cmt{Sample $\mathcal{X} \sim \mathbb{D}^m$ s.t. $\mathcal{X} \cap \calT = \emptyset$}\\
    $D_{A} \gets \bigcup_{x \in \mathcal{X}} \{(x, 0)\} $\\
    $D_{A} \gets D_{A} \cup \left(\bigcup_{x\in \calT} \{(x, 1)\}\right)$\\
    Train $h, W$ (Eq. \ref{equ:target}) from dataset $D_{A}$\\
    Initialize $\theta$\\
    $\theta \gets \theta \cup (h, W)$\\
    \textbf{Ret} $\theta$
    \end{flushleft}
  }
  {
  \begin{flushleft}
    \adversary{$\calA^{\mathbb{D},\calM}_{LDP}(t, G,\varepsilon)$}\\
    Extract $g_t$ as the gradient of the chosen neuron from $G$\\
    \textbf{Ret} $[g_t \neq 0]$
    \end{flushleft}
  }
 \end{framed}
 \caption{Attack Strategy of the Adversary $\calA_{LDP}$.}
\label{fig:strategy-ldp}
\end{figure}



\paragraph{Certified Guarantee of Success for AMI.} 
Now, we derive certified guarantees for the adversary (Fig. \ref{fig:strategy-ldp}) to be successful under $\varepsilon$-LDP protection. The AMI attack  $\calA^{\mathbb{D},\calM}_{LDP}$ is successful in determining the membership of the target sample $t$ if it can ensure that the chosen neuron is activated only by the LDP-preserving $\calM (t, \varepsilon)$. Following the expected output stability property in DP \cite{lecuyer2019certified}, in which the expected value of an $\varepsilon$-LDP algorithm with bounded output is not sensitive to small changes in the input,  the trained attack $\calA^{\mathbb{D},\calM}_{LDP}$ is certifiably robust to $\calM(\cdot, \varepsilon)$ if the following condition holds:
\begin{equation}\label{equ:cert-ldp}
\left \{
  \begin{aligned}
    & \mathbb{E} \big[ v (t) \big] > 0 \\
    &\mathbb{E} \big[ v (x)  \big]  \leq 0, \quad x \neq \calM (t, \varepsilon)
  \end{aligned} \right. 
\end{equation}
where $v (t)  =h \cdot \relu(W \calM (t, \varepsilon) ) $  and $v(x) = h \cdot \relu(W x) $ are the values of the chosen neuron, given the randomized target sample $\calM (t, \varepsilon)$ and any other data samples $x \neq \calM (t, \varepsilon)$, respectively.

However, due to the potentially complex nature of the post-noise
computation, we cannot precisely compute the expectations 
in Eq. (\ref{equ:cert-ldp}).  We therefore resort to Monte Carlo sampling to estimate the expectations $\hat{ \mathbb{E} }(\cdot) $.  This estimation is obtained by invoking $\calM (\cdot)$ multiple times with independent draws of the noise over the input. We denote $v_p(t)$ as the $p$  draws of $\calM (t, \varepsilon)$ from the target sample $t$ and $v_q(x)$ as the $q$ draws of $\calM (x, \varepsilon)$ from the  sample $x$. Then, we replace $\mathbb{E} \big[ v (t) \big]$ with $\hat{\mathbb{E}} \big[ v (t) \big] = \frac{1}{p} \sum_{p} v_p(t) $  and replace $\mathbb{E} \big[ v (x) \big]$ with $\hat{\mathbb{E}} \big[ v (x) \big] = \frac{1}{q} \sum_{q} v_q(x) $, where $p$ an $q$ are the number of invocations of $\calM (\cdot) $ for $t$ and $x$, respectively. 

The key idea is to simultaneously ensure that the lower bound $\hat{\mathbb{E}}^{lb} \big[ v (t) \big]$ is larger than $0$ and the upper bound  $\hat{\mathbb{E}}^{ub} \big[ v (x) \big]$ is smaller than or equal to $0$ with a broken probability $\delta$. That provides a certified guarantee for the Eq. (\ref{equ:cert-ldp}) to hold.
We compute $(1-\delta)$-confidence the lower bound $\hat{\mathbb{E}}^{lb} \big[ v (t) \big]$  and the upper bound  $\hat{\mathbb{E}}^{ub} \big[ v (x) \big]$ by using Hoeffding's inequality \cite{hoeffding}, as follows:
\begin{align}\label{equ:lb-ub}
  &  \hat{\mathbb{E}}^{lb} \big[ v (t) \big] \triangleq \hat{\mathbb{E}} \big[ v (t) \big] - \Range \big(v(t) \big) \sqrt{ - \frac{ \ln ( \delta ) }{ 2 p} } \\
   &  \hat{\mathbb{E}}^{ub} \big[ v (x) \big] \triangleq \hat{\mathbb{E}} \big[ v (x) \big] + \Range \big(v(x) \big) \sqrt{ - \frac{ \ln ( \delta ) }{ 2 q} }
   \label{equ:lb-ub2}
\end{align}
where $\Range\big(v(\cdot)\big)$ is the  range of $v(\cdot)$. 


By replacing the bounds in Eqs. (\ref{equ:lb-ub}) and (\ref{equ:lb-ub2}) to the expectations in Eq. (\ref{equ:cert-ldp}), we derive the certified guarantee so that Eq.~(\ref{equ:cert-ldp}) holds as in Theorem \ref{theorem:guarantee cond}.

\begin{theorem} (Certified Guarantee Condition) 
Suppose that clients in FL apply the LDP-preserving $\calM(\cdot, \varepsilon)$ to their local data. $\hat{\mathbb{E}}^{lb} \big[ v (t) \big]$ and $\hat{\mathbb{E}}^{ub}\big[ v (x)  \big]$, computed as in Eqs.~\ref{equ:lb-ub} and \ref{equ:lb-ub2}, are the $(1-\delta)$-confidence lower and upper bounds, respectively.
The AMI attack $\calA^{\mathbb{D},\calM}_{LDP}$  is successful in inferring the membership of the target sample $t$ in $\mathbb{D}$ if the following condition is satisfied: \begin{equation}
 \left \{
  \begin{aligned}
    & \hat{\mathbb{E}}^{lb} \big[ v (t) \big] > 0 \\
    &\hat{\mathbb{E}}^{ub}\big[ v (x)  \big]  \leq 0, \quad x \neq \calM (t, \varepsilon)
  \end{aligned} \right. 
\end{equation}
\label{theorem:guarantee cond}
\end{theorem}

Proof of Theorem \ref{theorem:guarantee cond} is in Appendix \ref{app: Proofs of Eqs}. 
At the attack time, we implement a \textit{certified guarantee of success} as a  search to return the minimal privacy budget $\varepsilon$ and  broken probability $\delta$ so that the condition in Theorem \ref{theorem:guarantee cond}  holds, as follows: 



\begin{corollary}
Given a well-trained model $f_{\theta}$ and the target sample $t$, 
the AMI attack $\calA^{\mathbb{D},\calM}_{LDP}$  is guaranteed to be successful up to the privacy budget $\varepsilon^*$ and the broken probability $\delta^*$ for which the condition in Theorem \ref{theorem:guarantee cond}  checks out:


{\small 
\begin{equation}
    (\varepsilon^*,~\delta^*){=}\arg\min_{\varepsilon, \delta}~\text{s.t.}~ \left \{
  \begin{aligned}
    & \hat{\mathbb{E}}^{lb} \big[  h \cdot \relu(W \mathcal{M}(t,\varepsilon)) \big] > 0 \\
 &\hat{\mathbb{E}}^{ub}\big[ h \cdot \relu(W x)  \big]  \leq 0, x {\neq} \calM (t, \varepsilon)
  \end{aligned} \right. 
\end{equation}
}

    


\end{corollary}

Since the mechanism $\calM(\cdot)$ and the privacy budget $\varepsilon$ are known to the adversary $\calA$, we only do a line search to find the minimal $\delta$ for a given $\varepsilon^*$. 






\section{EVALUATION}\label{sec:eval}
This section validates the effectiveness of our AMI attack by gauging its success rate. We particularly focus on evaluating  how well it performs under LDP protection. Our implementation of the attack is available at \url{https://github.com/trucndt/ami}.

\begin{figure*}[t]
    \centering
    \begin{subfigure}[b]{0.32\textwidth}
        \centering
        \includegraphics[width=\textwidth]{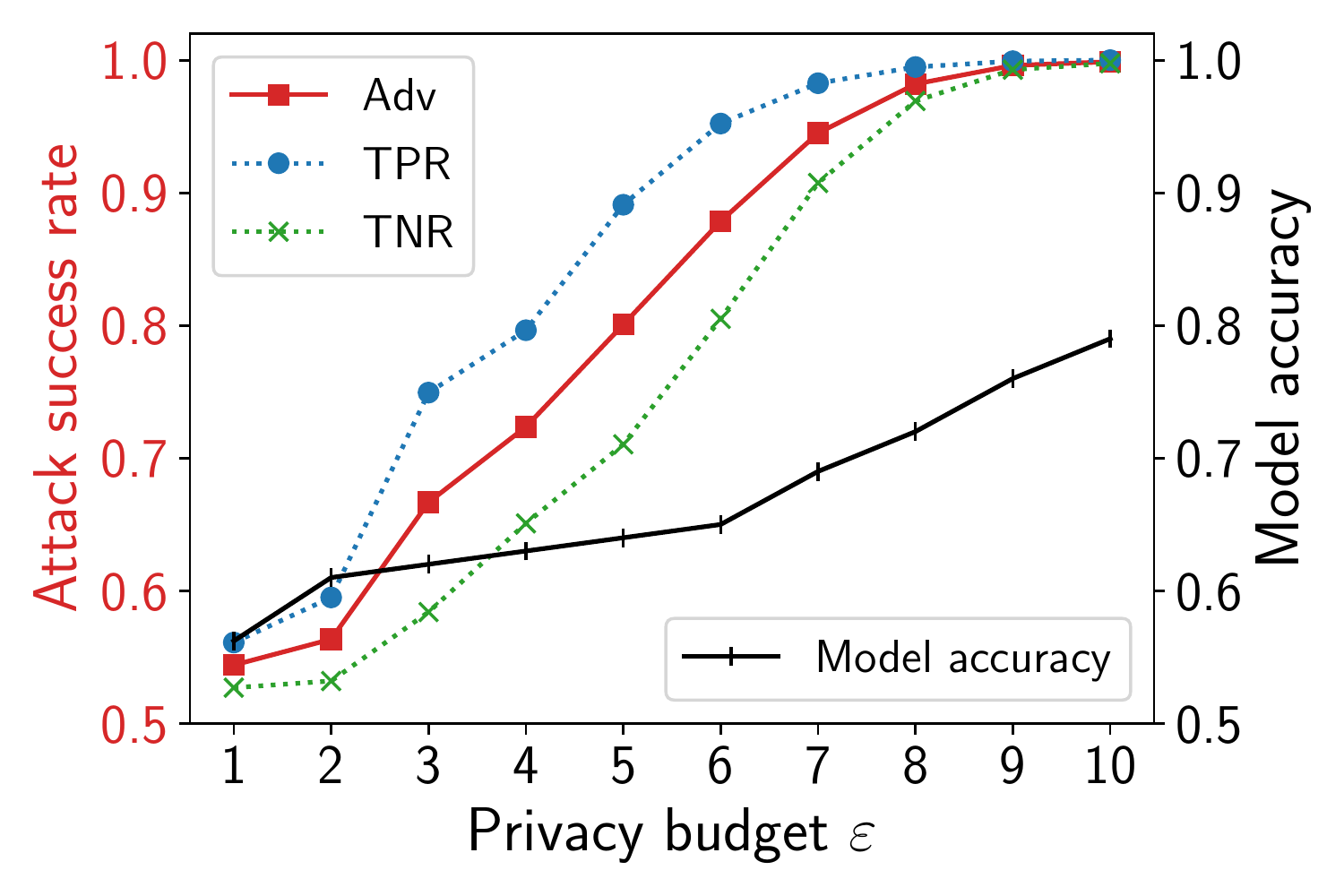}
        \caption{CelebA}
    \end{subfigure}
    \begin{subfigure}[b]{0.32\textwidth}
        \centering
        \includegraphics[width=\textwidth]{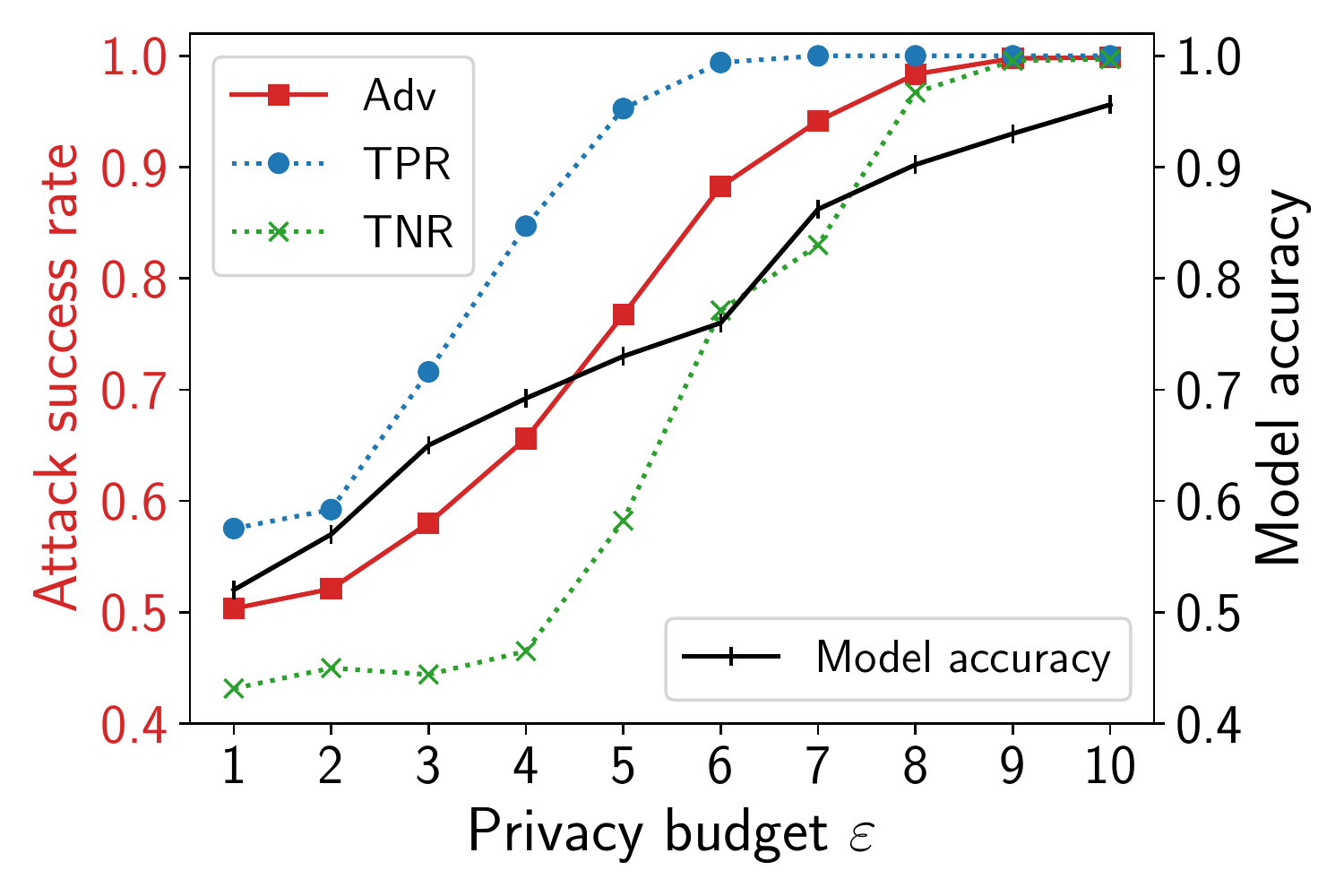}
        \caption{ImageNet}
    \end{subfigure}
    \begin{subfigure}[b]{0.32\textwidth}
        \centering
        \includegraphics[width=\textwidth]{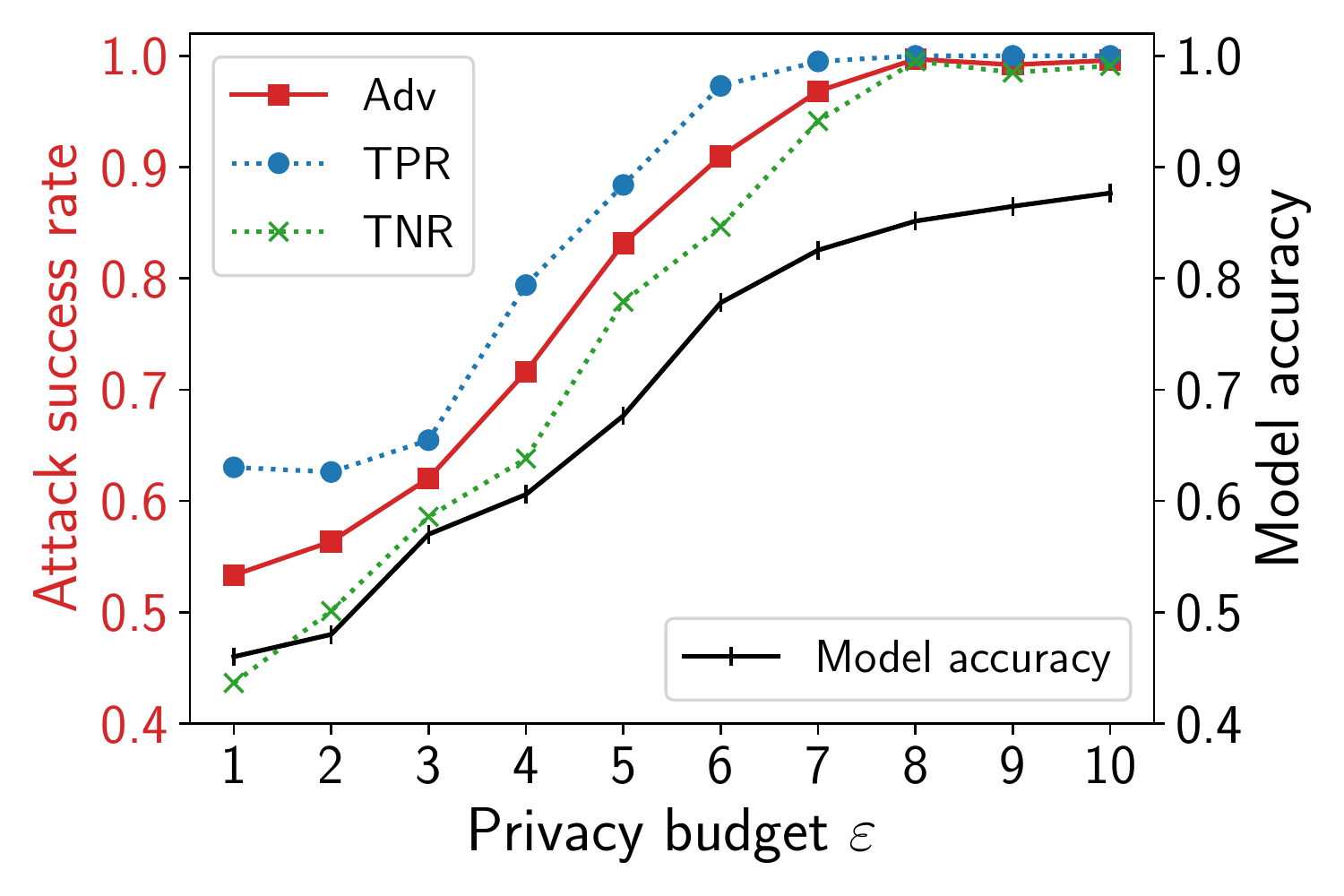}
        \caption{CIFAR-10}
    \end{subfigure}
    \caption{Attack success rate of AMI against an $\varepsilon$-LDP mechanism on CelebA, ImageNet, and CIFAR-10 datasets. The success rate is represented via the advantage (Adv), true positive rate (TPR), and true negative rate (TNR) according to Eq. \ref{equ:suc}. The baseline of random guessing is 0.5. The model accuracy illustrates the utility loss of the data when using LDP.}
    \label{fig:sucrate}
\end{figure*}


\begin{figure*}[t]
  \centering
\subfloat[$\varepsilon = 10$]{\label{fig:tsne-10}\includegraphics[scale=0.19]{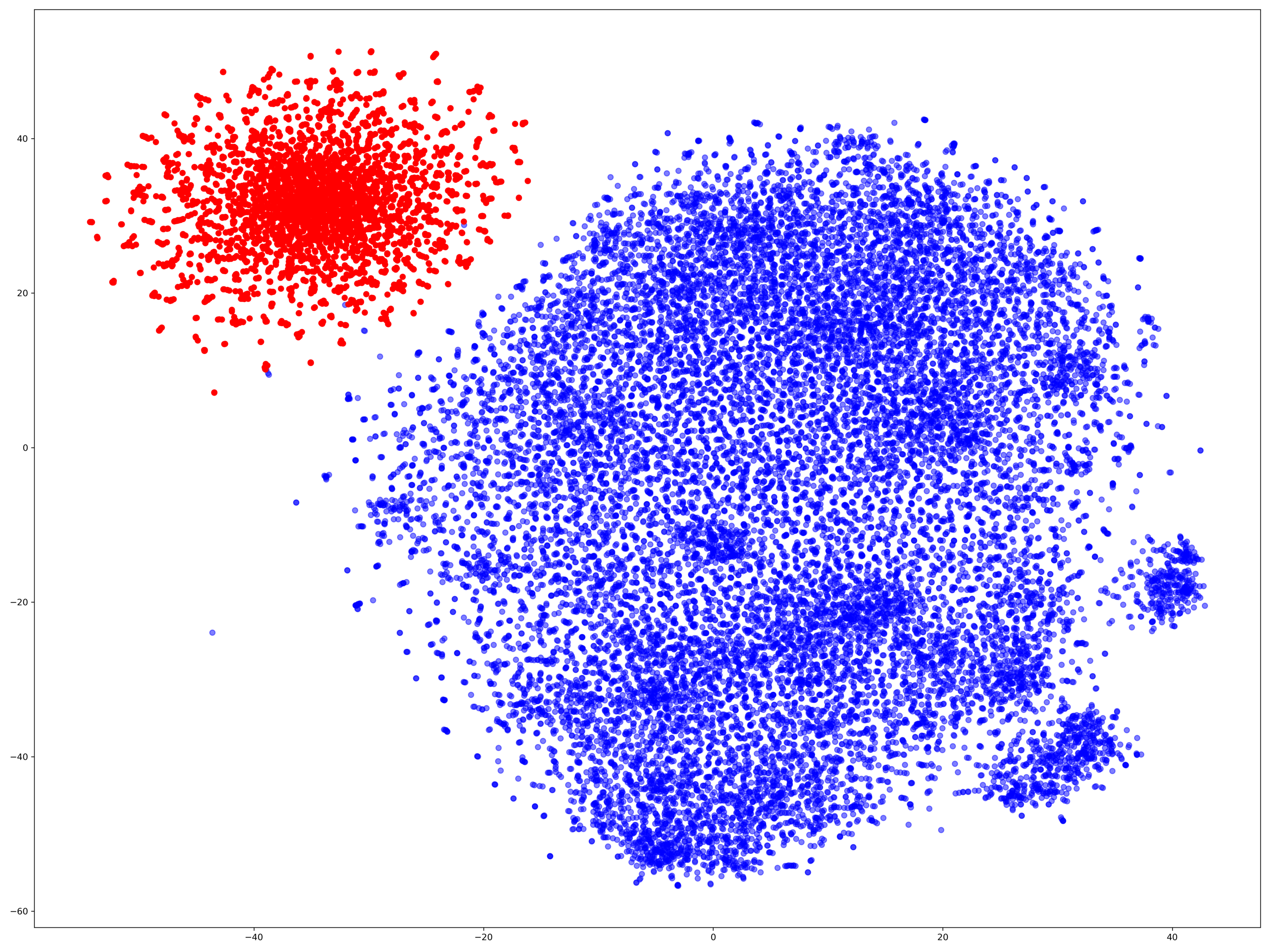}} 
\hspace{1cm}
\subfloat[$\varepsilon = 7$]{\includegraphics[scale=0.19]{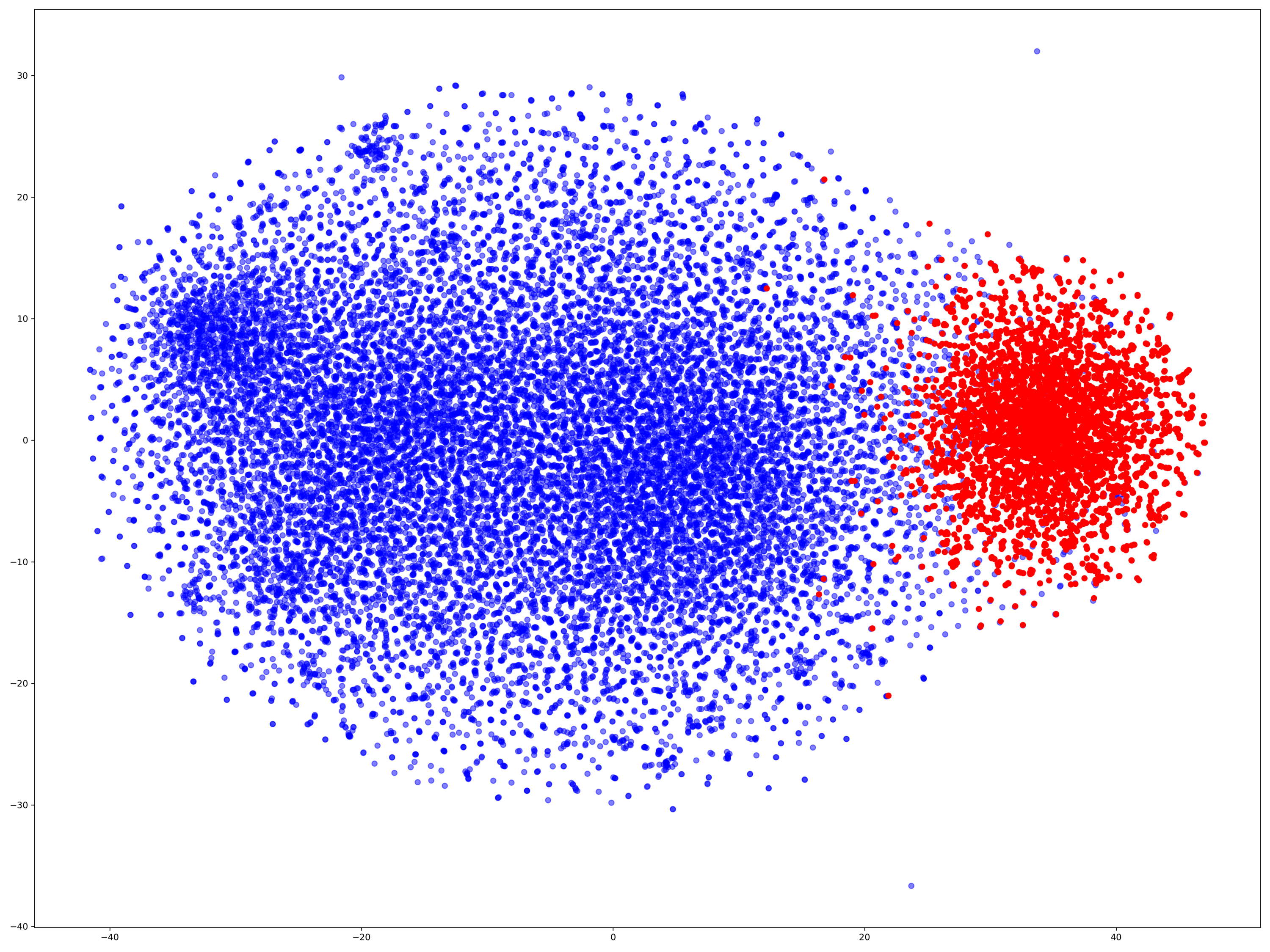}}\hspace{1cm}
\subfloat[$\varepsilon = 5$]{\label{fig:tsne-5} \includegraphics[scale=0.19]{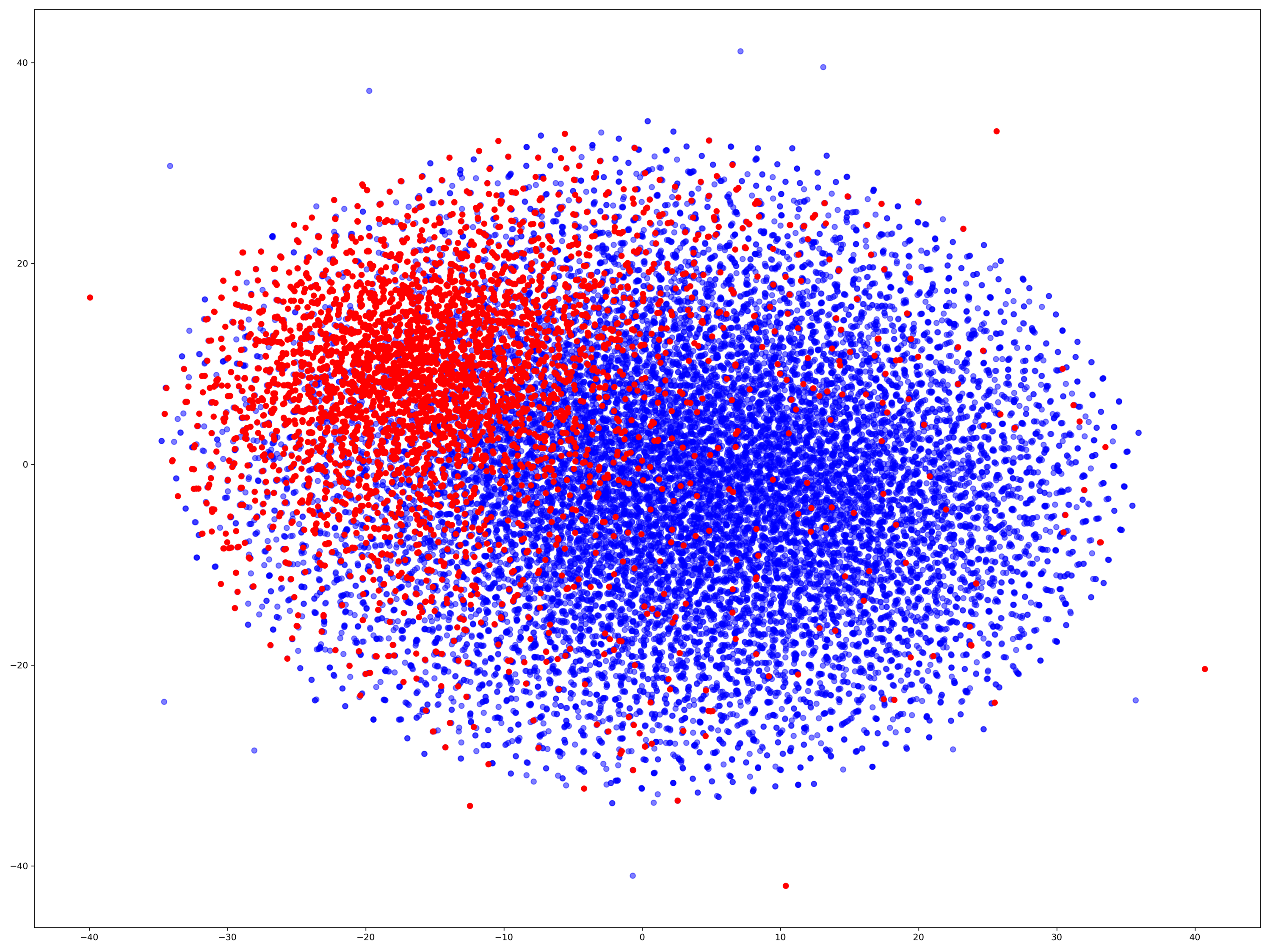}} 
      \caption{Visualizing the distribution of the target sample $t$ among other samples in the training set $\calD$ using t-SNE embeddings. The red dots denote the target sample $t$ and a multitude of its LDP noises $\calM(t, \varepsilon)$, while the blue dots denote other non-target samples. These data samples are obtained from the CelebA dataset.}
    \label{fig:tsne}
 \end{figure*}

\paragraph{Experimental Settings.} We evaluate AMI attack with three benchmark datasets, including CIFAR-10 \cite{krizhevsky2009learning}, ImageNet \cite{deng2009imagenet}, and CelebA \cite{liu2015faceattributes}. Each dataset includes a training set and a validation set. The training set is used to sample the local training set $\calD$ in the threat model (Figs. \ref{fig:threat-model} and \ref{fig:threat-model-ldp}), while the validation set is used as the sampled set $\mathcal{X}$ of the adversary. Our experiment follows the security game in Figs. \ref{fig:threat-model} and \ref{fig:threat-model-ldp}, and the success rate is calculated using Eq. (\ref{equ:suc}) after executing the game 10,000 times. The batch size $|\calD|$ is chosen to be 20 for the CelebA dataset according to its specification (i.e., 20 face images per person). As in \cite{fowl2021robbing,boenisch2021curious,geiping2020inverting}, we set $|\calD|$ to 64 and 100 for the ImageNet and CIFAR-10 datasets, respectively. We modify $r=1,000$ neurons in the first layer and 1 neuron in the second layer to carry out AMI attacks. Further details on the experimental settings can be found in Appendix \ref{app:exp}.


To realize $\calM$ in Fig.~\ref{fig:threat-model-ldp}, we use two different LDP mechanisms: OME \cite{lyu2020towards} and BitRand  \cite{lai2021bit}.
These mechanisms add LDP noises to the embeddings of data samples. Such embeddings are obtained via a pre-trained Resnet-18 model which results in feature vectors of 512 dimensions \cite{he2016deep}. We show the results when using BitRand in this section and refer the readers to Appendix \ref{app:exp} for the results on OME. A background on BitRand and OME is provided in Appendix \ref{app:ldp}.

\paragraph{Attack Performance without LDP.} 

In three datasets, our attack achieves near 100\% success rate. The key reason behind this impressive success rate is that our attack strategy can easily train the chosen neuron to satisfy the attack objective, i.e., Eq. (\ref{equ:cond-nonlinear}). 
Intuitively, the problem formulation in Eq. (\ref{equ:cond-nonlinear}) is equivalent to finding a decision boundary over-fitting to $t$ in a way that can distinguish $t$ against all other samples. As a result, increasing the number of neurons in the first layer ($r$) 
helps improve the attack performance, as it makes the model more over-fitting. In our experiments, we can achieve a 100\% success rate with as few as 5 neurons ($r=5$) in the first layer. We refer the readers to Appendix \ref{app:exp} for further analysis of this scenario as we focus the rest of this section on evaluating the attack under LDP.

\paragraph{Attack Performance under LDP.} Fig. \ref{fig:sucrate} shows that our attack introduces severe privacy risk to clients' local training data through strong attack success rates under LDP protection. 
Large privacy budgets $\varepsilon$ (e.g., $\varepsilon \geq 5$) does minimal to defend against our AMI attack.
Across all three datasets, the model accuracy on the legitimate classification task remains acceptable given $\varepsilon \geq 5$. However, our attack imposes a severely high success rate ($\geq 0.77$), which approaches a near perfect success rate of $0.99$ with $\varepsilon \geq 9$. When we reduce the privacy budget ($\varepsilon \in [3, 4]$), our attack still maintains a success rate of at least 0.67, 0.58, and 0.62, on CelebA, ImageNet, and CIFAR-10, respectively. 
With very low $\varepsilon$ ($\varepsilon \in [1, 2]$), the model accuracy is severely damaged.


Furthermore, Fig. \ref{fig:sucrate} depicts the TPR and TNR of our attack. Recall that TPR denotes how well the attack detects the presence of the target sample $t$ in the training data $\calD$, and TNR measures the ability to detect the absence of $t$. From the result, we can see that our attack has high TPR across all scenarios, which means it is sensitive to detecting the case where $t\in D$. Moreover, our TNR is greater than 0.5 indicating the capability of discerning the absence of $t$ in the training data (except for the ImageNet dataset at $\varepsilon \leq 4$). 


\begin{figure*}[t]
  \centering
\subfloat[CelebA]{\includegraphics[scale=0.32]{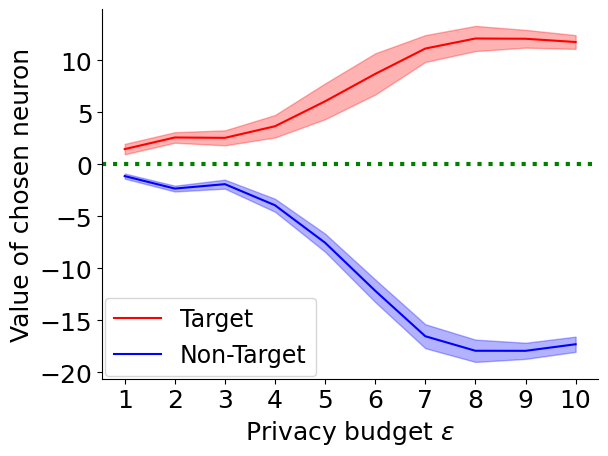}} 
\hspace{1cm}
\subfloat[ImageNet]{\includegraphics[scale=0.32]{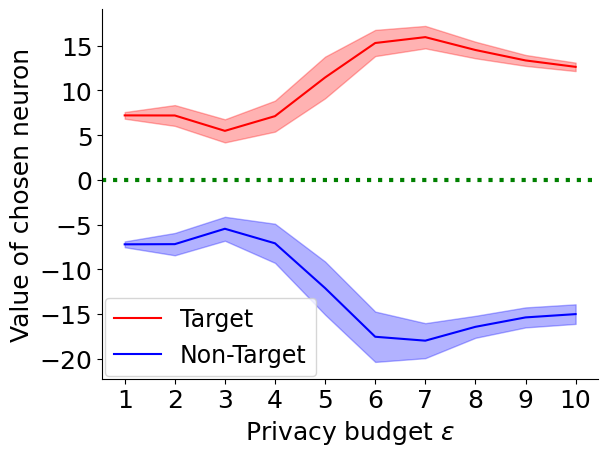}}\hspace{1cm}
\subfloat[CIFAR-10]{\includegraphics[scale=0.32]{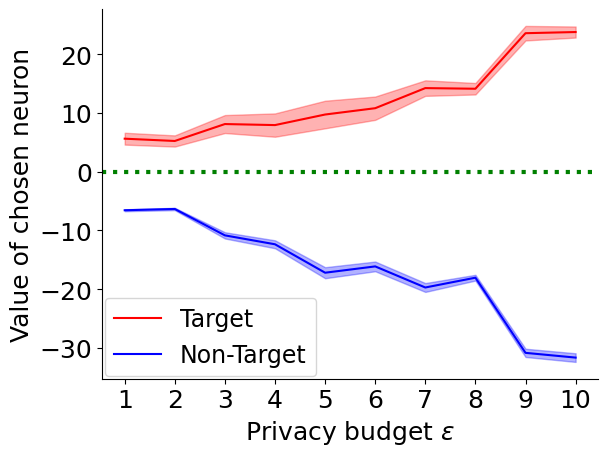}} 
      \caption{Certified guarantee of success for $\varepsilon \in [1,10]$: Expectation (solid lines), upper-bound, and lower-bound
(shaded areas surrounding the expectation) of the values of the chosen neuron. The larger the gap between the lower bound
of the noisy target samples and the upper bound of the noisy non-target samples, the higher success rate the AMI achieves.}
    \label{fig:bound}
 \end{figure*}

\paragraph{Training the Chosen Neuron under LDP.}
Training the chosen neuron is equivalent to determining a decision boundary that can distinguish the target sample (and its randomized variants) from any other samples. Fig. \ref{fig:tsne} visualizes how the samples in the training set $\calD$ are distributed using t-SNE \cite{van2008visualizing}. 
At $\varepsilon=10$, Fig. \ref{fig:tsne-10} shows that the t-SNE algorithm is able to group together the target sample $t$ and its randomized variants. This is because the LDP mechanism imposes a small amount of noise such that $t$ and its randomized variants $\calM(t, \varepsilon)$ closely resemble one another. Hence, t-SNE models these by nearby points. Therefore, it is easy for our attack to train a  neuron that can distinguish $\calM(t, \varepsilon)$ from other samples, resulting in an attack success rate of about 0.99 as shown in Fig. \ref{fig:sucrate}.

At $\varepsilon=5$, Fig. \ref{fig:tsne-5} shows that $\calM(t, \varepsilon)$ blends into other samples, meaning that t-SNE is unable to group together the randomized variants of $t$ as in the previous Fig. \ref{fig:tsne-10}. This is because the mechanism $\calM(\cdot, \varepsilon)$ imposes a high amount of noise at $\varepsilon=5$, so that all randomized variants $\calM(t,\varepsilon)$ no longer closely resemble one another. This makes the task of finding the decision boundary between $\calM(t,\varepsilon)$ and other samples more difficult. Nevertheless, our AMI attack can still attain a success rate of 0.80 (Fig.~ \ref{fig:sucrate}).

\paragraph{Certified Guarantee of Success.}
 Given a privacy budget $\varepsilon \in [1,10]$, 
in order to check the certified guanrantee conditions as in Theorem \ref{theorem:guarantee cond}, we obtain the 
the lower and upper bounds $\hat{\mathbb{E}}^{lb} \big[ v (t) \big] $ and $\hat{\mathbb{E}}^{ub}\big[ v (x)  \big]$ (Eqs.~\ref{equ:lb-ub} and \ref{equ:lb-ub2})  by using $4,000$ $\varepsilon$-LDP target samples $\calM (t, \varepsilon)$ and all $\varepsilon$-LDP non-target samples $\calM (x, \varepsilon)$ from the validation set of each dataset. Here we use BitRand \cite{lai2021bit} as the   $\calM(\cdot, \varepsilon)$ mechanism for the embeddings since BitRand is designed and well-suited for randomizing the embeddings. 


Fig.~\ref{fig:bound} shows the certified guarantee of
success for the CelebA, ImageNet, and CIFAR-10 datasets. We can derive certified guarantee of success for our AMI attack given $\varepsilon \geq 1$ with a small broken probability $10^{-8}$. For rigorous privacy budgets, e.g., $\varepsilon \leq 3$, the output of the chosen neuron for both the target and non-target samples approaches the borderline associated with $v(\cdot) = 0$ (i.e., the dotted green lines), indicating a higher chance for AMI attacks to be failed given a broken probability $\delta$. When the privacy budget $\varepsilon$ increases, the output of the chosen neuron fits well with the attack objective. As a result,  the expected value of the chosen neuron departs from the borderline, i.e., more positive given the target samples (i.e., the solid red lines) and more negative given the non-target samples (i.e., the solid blue lines). That implies a better attack success rate. Also, we observed that the overlapping area between the two distributions of the target samples and the non-target samples reduce significantly, which is consistent with our certified guarantee of success (Fig.~\ref{fig:histogram-neuron-value}, Appendix \ref{app:exp}) and our empirical results in Fig.~\ref{fig:sucrate}. 

\section{RELATED WORK}\label{sec:related}

Membership inference (MI) is one of the most fundamental privacy problems in machine learning \cite{carlini2022membership}. Several research has been carried out to convey the practical consequences of MI attacks \cite{backes2016membership,pyrgelis2017knock} and analyze the models' vulnerability to MI \cite{carlini2019secret,song2021systematic}. Along this direction, multiple MI attacks have been proposed in which the attacker only queries the model or observes its parameters to conduct the attacks \cite{shokri2017membership,salem2018ml,carlini2022membership}. Such attacks can be straightforwardly adapted to FL in which the central server is a passive adversary who tries to infer the membership information of clients' private data via inspecting their local models' parameters \cite{nguyen2022preserving,melis2019exploiting}.
Recently, an AMI attack in FL proposed by \cite{DBLP:conf/sp/NasrSH19} considers a dishonest server that can interfere with the FL training protocol. However, this attack must be repeated in multiple training iterations to attain a high success rate. Furthermore, LDP has been shown to be an effective defense against these attacks \cite{rahman2018membership,10.1007/978-3-030-81242-3_2,GU2022103201}.

Our work proposes a new AMI attack in FL where the dishonest server can maliciously modify the model weights to its advantage. We have proposed a strategy that results in minimal modifications to the model and can be executed in only one training iteration. More importantly, our attack can maintain a strong success rate even when the clients' data are protected by an LDP mechanism.

\section{CONCLUSION AND DISCUSSION}\label{sec:conclude}

In this paper, we have introduced a formal threat model for our AMI attack with dishonest FL servers, showing a more realistic privacy threat. Accordingly, we have proposed a new active membership inference (AMI) attack, exploiting the correlation among data features through a non-linear decision boundary. AMI attacks can achieve high success rates even under LDP protection, confirmed by both theoretical analysis and experimental evaluations. From this attack, our research has demonstrated that current implementations of FL provide virtually no privacy protection for clients.

With such a strong AMI attack, our future work would focus on the defenses. We discuss some challenges in devising such a solution as follows.

\begin{figure}
    \centering
    \includegraphics[width=1\linewidth]{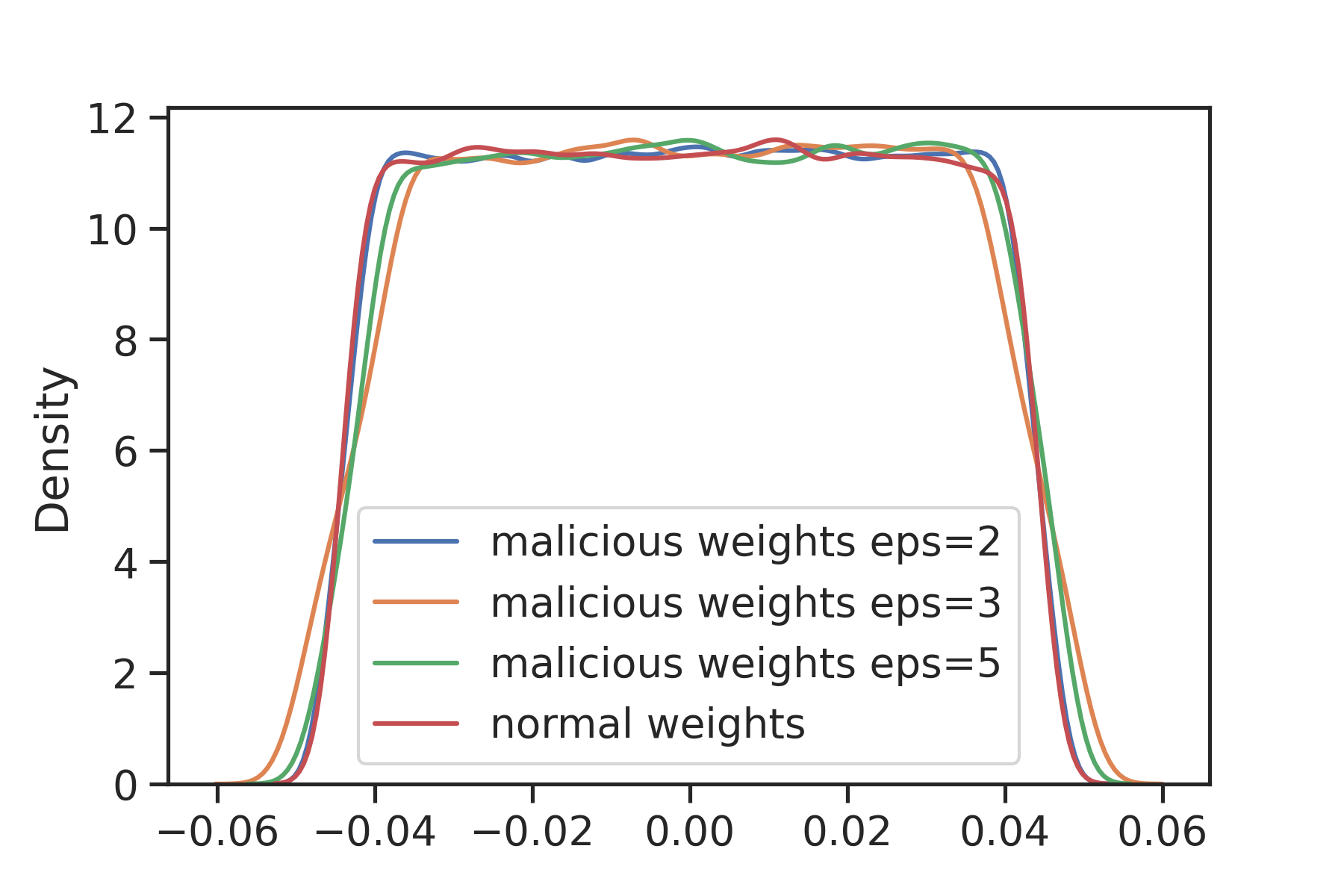}
    \caption{The kernel density estimation (KDE) of malicious and normal weights.}
    \label{fig:dist-cmp}
\end{figure}
\textbf{Noisy gradients with DPSGD.} A potential defense against our attack is to let clients add DP noise to their gradients using DPSGD \cite{abadi2016deep} before sending them to the server, hindering the attacker from knowing the true value of the chosen neuron's gradient. However, recent work \cite{boenisch2021curious,tramer2020differentially} suggests that using DPSGD makes it impossible to train a good model for datasets like CIFAR-10 or ImageNet. Furthermore, even if DPSGD is used, the attacker can still circumvent it by aggregating the noisy gradients over multiple FL iterations and potentially cancelling out the added DP noise. A detailed analysis on this can be found in Appendix \ref{sec:def}.

\textbf{Detecting Malicious Weights is Challenging.} The hardness in detecting malicious weights can be evaluated by examining the difference between malicious weights and normal weights (i.e., the weights that could be obtained from an honest server). Fig. \ref{fig:dist} (Appendix) shows the distribution of the normal weights together with the distribution of the malicious weights when attacking under LDP protection at $\varepsilon=2,3,$ and $5$, and they largely resemble one another. Fig. \ref{fig:dist-cmp} shows the 4 distributions using their respective kernel density estimation (KDE), which is used to visualize the shape of a data distribution, and represent the data using a continuous probability density curve. We can see that the malicious weights do not result in any abnormal distribution, making it indistinguishable from normal weights. This implies that, by observing the distribution of model weights, it is infeasible to determine whether the model weights have been modified maliciously by our attack.

\subsubsection*{Acknowledgements}
This material is based upon work supported by the National Science Foundation under grants CNS-1935928, CNS-1935923, and CNS-2140477.

\bibliography{iclr2023_conference}
\bibliographystyle{apalike}

\clearpage

\appendix
\onecolumn

\section{INFEASIBILITY OF LINEARITY FOR AMI}\label{app:linear}
Suppose that, given the target data sample $t\in \mathbb{R}^d$, there exists a $W$ that can satisfy Eq. (\ref{equ:linear}) for all $x\neq t$. We choose $x_1 = t_1 + c$ and $x_i = t_i$ for $i > 1$ and $c > 0$. Denoting $w \equiv W_i$, from the second condition in Eq. (\ref{equ:linear}), we have that:

\begin{equation}\label{equ:contradict1}
    w_1c + \sum_{j=1}^d w_j t_j \leq 0 \Longrightarrow -w_1 c \geq \sum_{j=1}^d w_j t_j
\end{equation}

Likewise, choosing $x'_1 = t_1 - c$ and $x'_i = t_i$ for $i > 1$, from the second condition in Eq. (\ref{equ:linear}), we have:

\begin{equation} \label{equ:contradict2}
    -w_1c + \sum_{j=1}^c w_j t_j \leq 0 \Longrightarrow w_1 c \geq \sum_{j=1}^d w_j t_j
\end{equation}

As $\sum_{j = 1}^d w_j t_j > 0$ by the first condition in Eq. (\ref{equ:linear}), the two equations (\ref{equ:contradict1}) and (\ref{equ:contradict2}) contradicts one another. Therefore, there exists no $W$ that can satisfy Eq. (\ref{equ:linear}) for all $x\neq t$.

\section{PROOF OF THEOREM \ref{theorem:guarantee cond} } \label{app: Proofs of Eqs}

Given the target sample $t$ and any data samples  $x \neq \calM (t, \varepsilon)$, 
the AMI attack  $\calA^{\mathbb{D},\calM}_{LDP}$ is successful in determining the membership of  $t$ if it can ensure that the chosen neuron is activated only by the LDP-preserving $\calM (t, \varepsilon)$. Following the expected output stability property in DP \cite{lecuyer2019certified}, in which the expected value of an $\varepsilon$-LDP algorithm with bounded output is not sensitive to small changes in the input,  the trained attack $\calA^{\mathbb{D},\calM}_{LDP}$ is certifiably robust to $\calM(\cdot, \varepsilon)$ if the following condition holds:
\begin{equation}\label{app-equ:cert-ldp}
\left \{
  \begin{aligned}
    & \mathbb{E} \big[ v (t) \big] > 0 \\
    &\mathbb{E} \big[ v (x)  \big]  \leq 0, \quad x \neq \calM (t, \varepsilon)
  \end{aligned} \right. 
\end{equation}
where $v (t)  =h \cdot \relu(W \calM (t, \varepsilon) ) $  and $v(x) = h \cdot \relu(W x) $ are the values of the chosen neuron, given the randomized target sample $\calM (t, \varepsilon)$ and any other data samples $x \neq \calM (t, \varepsilon)$, respectively.

However, due to the potentially complex nature of the post-noise
computation, we cannot precisely compute the expectations 
in Eq.~\ref{equ:cert-ldp}.  We therefore resort to Monte Carlo sampling to estimate the expectations $\hat{ \mathbb{E} }(\cdot) $.  This estimation is obtained by invoking $\calM (\cdot)$ multiple times with independent draws of the noise over the input. We denote $v_p(t)$ as the $p$  draws of $\calM (t, \varepsilon)$ from the target sample $t$ and $v_q(x)$ as the $q$ draws of $\calM (x, \varepsilon)$ from the  sample $x$.

Denoting $\Range\big(v(\cdot)\big)$ as the  range of $v(\cdot)$,  $v (\cdot)  \in \Range \big( v (\cdot) \big)$. In other words, $v (\cdot)$ is bounded in $\Range \big( v (\cdot) \big)$. Given a broken probability $\delta$, using Hoeffding's inequality, with $t \geq 0$ we have:

\begin{align}
 \nonumber   P \Big( (\frac{1}{p} \sum_{i=1}^p v(t) ) - \mathbb{E} [v(t)] \geq t \Big) &
= P \Big( \frac{1}{p}  \sum_{i=1}^p \big( v(t)  - \mathbb{E} [v(t)] \big) \geq t \Big)\\
 & \leq  \exp \Big( - \frac{ 2 p t^2 }{ \frac{1}{p} \sum_{i=1}^p \Range \big( v(t) \big)^2 } \Big) =  \exp \Big( - \frac{ 2 p t^2 }{  \Range \big( v(t) \big)^2 } \Big) 
  \label{eq:proof}
\end{align}

As mentioned in Section \ref{sec:ldp}, we replace $\mathbb{E} [v(t)]$ in Eq.~\ref{eq:proof} with $ \hat{\mathbb{E}} [v(t)]$. Given a broken probability $\delta$, we have:

\begin{align}
 \exp \Big( - \frac{ 2 p t^2 }{  \Range ( v(t) )^2 } \Big)  = \delta 
  \Leftrightarrow t = \Range \big(v(t) \big) \sqrt{ - \frac{ \ln ( \delta ) }{ 2 p} }
   \label{eq:t-target}
\end{align}

Similarly, with the non-target samples, we have: 
\begin{equation}
  t = \Range \big(v(x) \big) \sqrt{ - \frac{ \ln ( \delta ) }{ 2 q} }  
  \label{eq:t-nontarget}
\end{equation}

By leveraging the Monte Carlo sampling for the expectation estimation,  we can replace $\mathbb{E} \big[ v (t) \big]$ with $\hat{\mathbb{E}} \big[ v (t) \big] = \frac{1}{p} \sum_{p} v_p(t) $  and replace $\mathbb{E} \big[ v (x) \big]$ with $\hat{\mathbb{E}} \big[ v (x) \big] = \frac{1}{q} \sum_{q} v_q(x) $, where $p$ an $q$ are the number of invocations of $\calM (\cdot) $ for $t$ and $x$, respectively. 

The key idea is to simultaneously ensure that the lower bound $\hat{\mathbb{E}}^{lb} \big[ v (t) \big]$ is larger than $0$ and the upper bound  $\hat{\mathbb{E}}^{ub} \big[ v (x) \big]$ is smaller than or equal to $0$ with a broken probability $\delta$. That provides a certified guarantee for the Eq. (\ref{equ:cert-ldp}) to hold.
From Eqs.~\ref{eq:t-target} and \ref{eq:t-nontarget}, we can compute $(1-\delta)$-confidence the lower bound $\hat{\mathbb{E}}^{lb} \big[ v (t) \big]$  and the upper bound  $\hat{\mathbb{E}}^{ub} \big[ v (x) \big]$, as follows:

\begin{align}\label{app-equ:lb-ub}
  &  \hat{\mathbb{E}}^{lb} \big[ v (t) \big] \triangleq \hat{\mathbb{E}} \big[ v (t) \big] - \Range \big(v(t) \big) \sqrt{ - \frac{ \ln ( \delta ) }{ 2 p} } \\
   &  \hat{\mathbb{E}}^{ub} \big[ v (x) \big] \triangleq \hat{\mathbb{E}} \big[ v (x) \big] + \Range \big(v(x) \big) \sqrt{ - \frac{ \ln ( \delta ) }{ 2 q} }
   \label{app-equ:lb-ub2}
\end{align}

By replacing the bounds in Eqs.~\ref{app-equ:lb-ub} and \ref{app-equ:lb-ub2} to the expectations in Eq.~\ref{app-equ:cert-ldp}, we we derive the certified guarantee so that Eq. (\ref{app-equ:cert-ldp}) holds. In other words, The AMI attack $\calA^{\mathbb{D},\calM}_{LDP}$  is successful in inferring the membership of the target sample $t$ in $\mathbb{D}$ if the following condition is satisfied: 

\begin{equation}
 \left \{
  \begin{aligned}
    & \hat{\mathbb{E}}^{lb} \big[ v (t) \big] > 0 \\
    &\hat{\mathbb{E}}^{ub}\big[ v (x)  \big]  \leq 0, \quad x \neq \calM (t, \varepsilon)
  \end{aligned} \right. 
\end{equation}

Consequently, Theorem \ref{theorem:guarantee cond}   holds.

\section{OME \cite{lyu2020towards} AND BITRAND \cite{lai2021bit}   }\label{app:ldp}

Apart from applying LDP-preserving mechanisms in real values of inputs or gradients  \cite{warner1965randomized,zhao2020local,wang2019collecting}, there is a line of work introducing LDP-preserving mechanisms to inputs or embedding features \cite{lai2021bit, lyu2020towards, arachchige2019local}. In these mechanisms, they encode the original data or embedding features into binary vectors, then apply the LDP mechanisms on top of the binary vectors, before training the local models.

In OME, each bit $i$ is randomized differently depending on whether it is the odd or even bit or it is bit $0$ or $1$, as follows:  
\begin{align}
   \forall i \in [0, rl-1]:   P(v'_x(i)=1) = \left \{
  \begin{aligned}
    &p_{1X} = \frac{\alpha}{ 1+\alpha}, \text{ if } i \in 2j, v_x(i) = 1 \\ 
    &p_{2X} = \frac{1}{1+\alpha^3},   \text{ if } i \in 2j+1, v_x(i) = 1  \\
    &q_X = \frac{1}{1+ \alpha \exp (\frac{\varepsilon}{rl})},  \text{ if }\ v_x(i) = 0 
  \end{aligned} \right.
   \label{ome_prob}
\end{align}
where $\varepsilon$ the total privacy budget. 

This mechanism is similar to the Utility enhancing randomization (UER) mechanism (Theorem III.4 \cite{arachchige2019local}). As shown in \cite{arachchige2019local, lyu2020towards}, model accuracy is almost constant although  $\varepsilon$  is changed. 

However, existing LDP mechanisms suffer from the curse of privacy composition in which excessive privacy budgets are consumed proportionally to the large dimensions of input or embedded features \cite{arachchige2019local}, gradients \cite{zhao2020local,wang2019collecting}, and training rounds \cite{zhao2020local,wang2019collecting}, causing loose privacy protection or inferior model accuracy \cite{wagh2021dp}. 


To mitigate the curse of privacy composition and to optimize the trade-off among privacy and model utility, \cite{lai2021bit} introduce bit-aware term $\frac{i \% l}{l}$ and a temperature $\alpha$ for better control of the randomization probabilities. In \textsc{BitRand}, the randomization probabilities are adaptively randomized
such that \textit{``bits with a more substantial impact''} on model utility will have \textit{''smaller randomization
probabilities (less noisy)''} and vice-versa under the same privacy budget, as follows:
\begin{align}
  \forall i \in [0, rl-1]:  P(v'_x(i)=1) = \left \{
  \begin{aligned}
    &p_X = \frac{1}{1+ \alpha \exp (\frac{i \% l}{l}\varepsilon)}, \text{ if}\ v_x(i) = 1 \\
    &q_X = \frac{\alpha \exp (\frac{i \% l}{l}\varepsilon)}{1+ \alpha \exp (\frac{i \% l}{l}\varepsilon)}, \text{ if}\ v_x(i) = 0 
  \end{aligned} \right.
  \label{f-RRRP} 
\end{align}
where $v_x(i) \in \{0, 1\}$ is the value of $v_x$ at the bit $i$, $v'_x$ is the perturbed vector created by randomizing all the bits in $v_x$, $\varepsilon$ is a privacy budget, and $\alpha$ is a parameter bounded with $ 0 < \alpha \leq \sqrt{\frac{\varepsilon + rl}{2r \sum_{i=0}^{l-1} \exp (2\frac{\varepsilon }{l}i\%l)}}$. The bit-aware term $\frac{i \% l}{l}$ to indicate the location of bit $i$, which is associated with the sensitivity of the bit at that location, in its $l$-bit binary encoded vector among $rl$ concatenated binary bits.


\section{ADDITIONAL EXPERIMENTS}\label{app:exp} \label{app:add exp}
This section provides more details on the experimental settings, and presents additional experiments.

\paragraph{Settings.} Our experiments in this paper are implemented using Python 3.8 and conducted on a single GPU-assisted compute node that is installed with a Linux 64-bit operating system. The allocated resources include 8 CPU cores (AMD EPYC 7742 model) with 2 threads per core, and 60GB of RAM. The node is also equipped with 8 GPUs (NVIDIA DGX A100 SuperPod model), with 80GB of memory per GPU.

\begin{figure*}[t]
    \centering
    \begin{subfigure}[b]{0.32\textwidth}
        \centering
        \includegraphics[width=\textwidth]{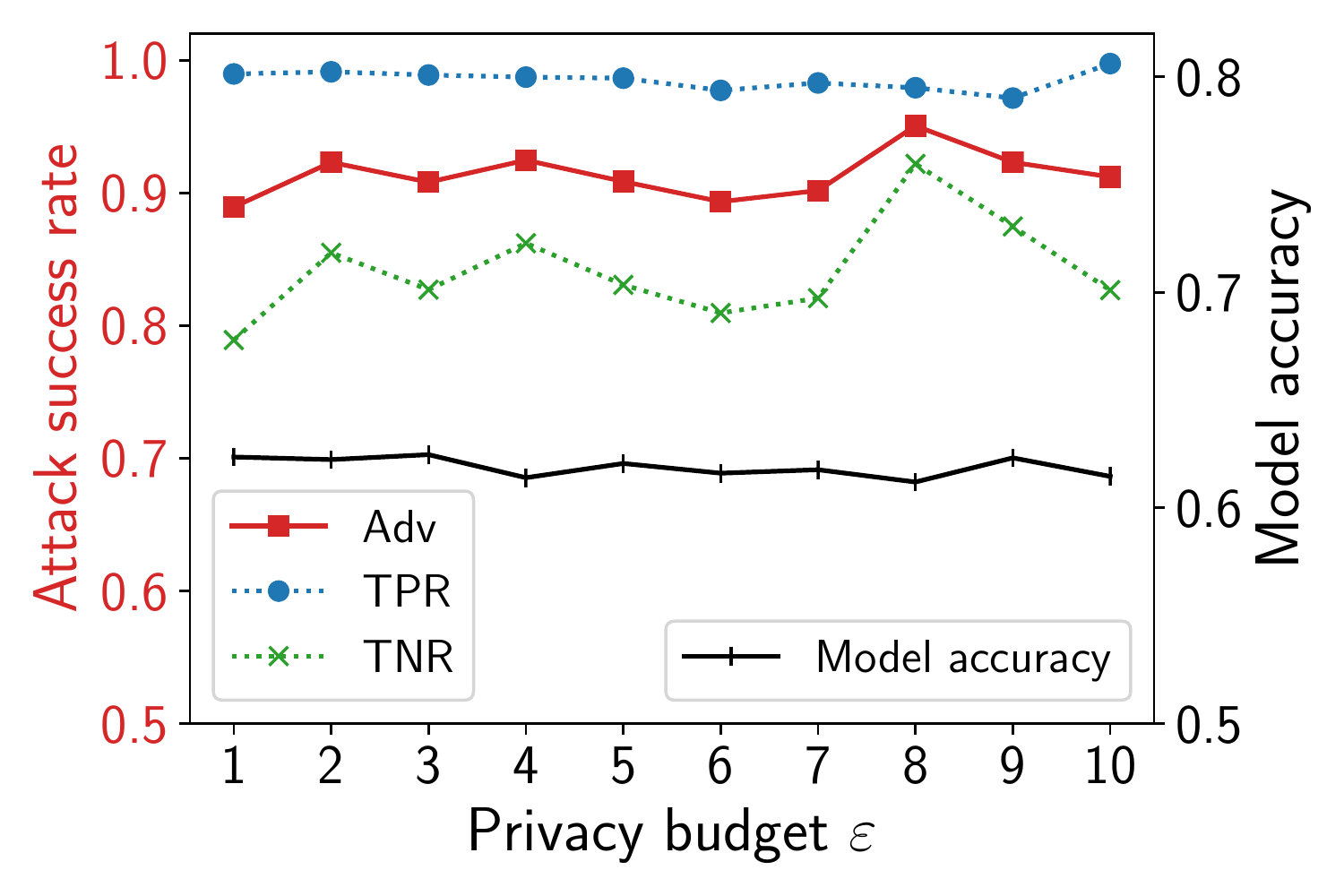}
        \caption{CelebA}
    \end{subfigure}
    \begin{subfigure}[b]{0.32\textwidth}
        \centering
        \includegraphics[width=\textwidth]{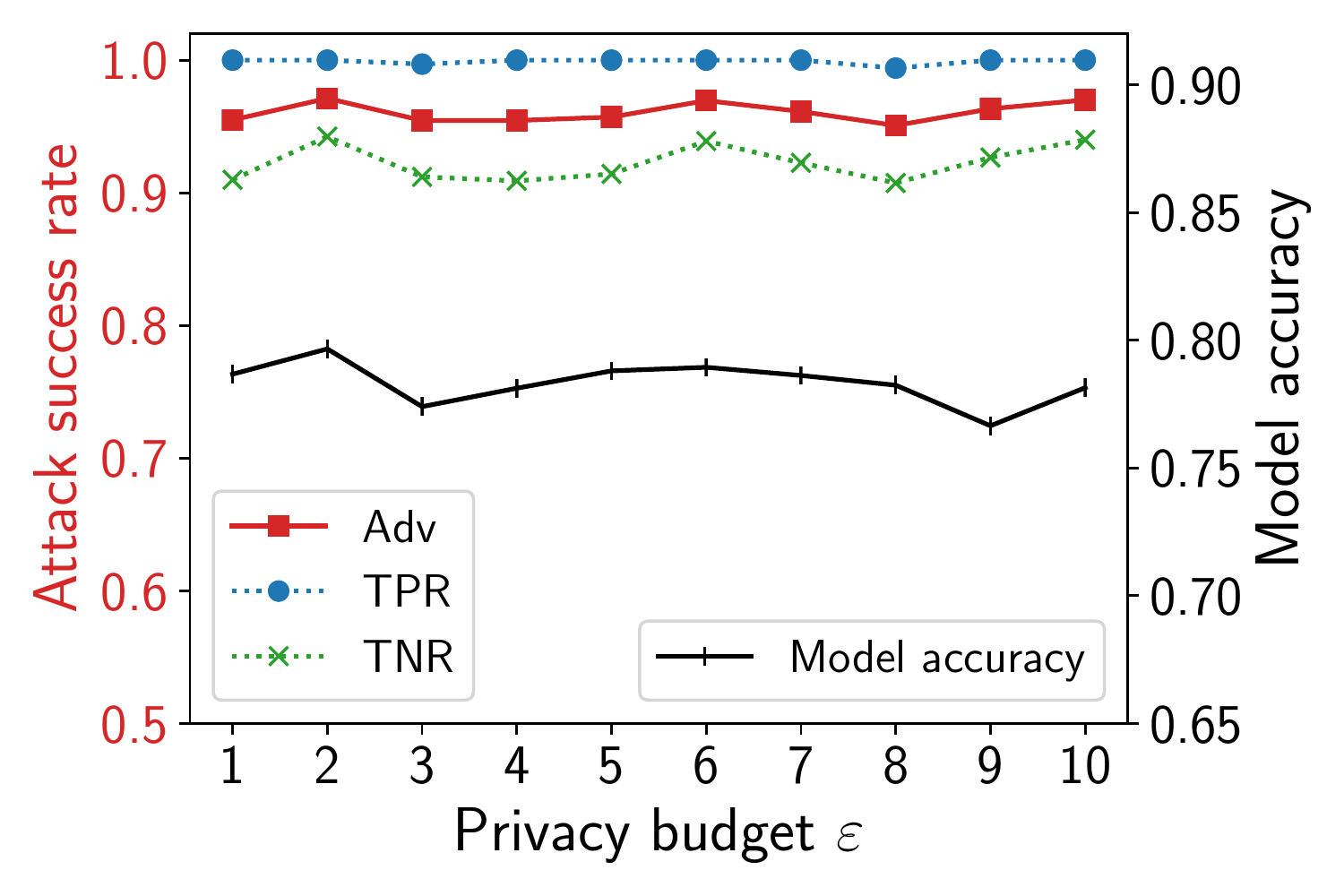}
        \caption{ImageNet}
    \end{subfigure}
    \begin{subfigure}[b]{0.32\textwidth}
        \centering
        \includegraphics[width=\textwidth]{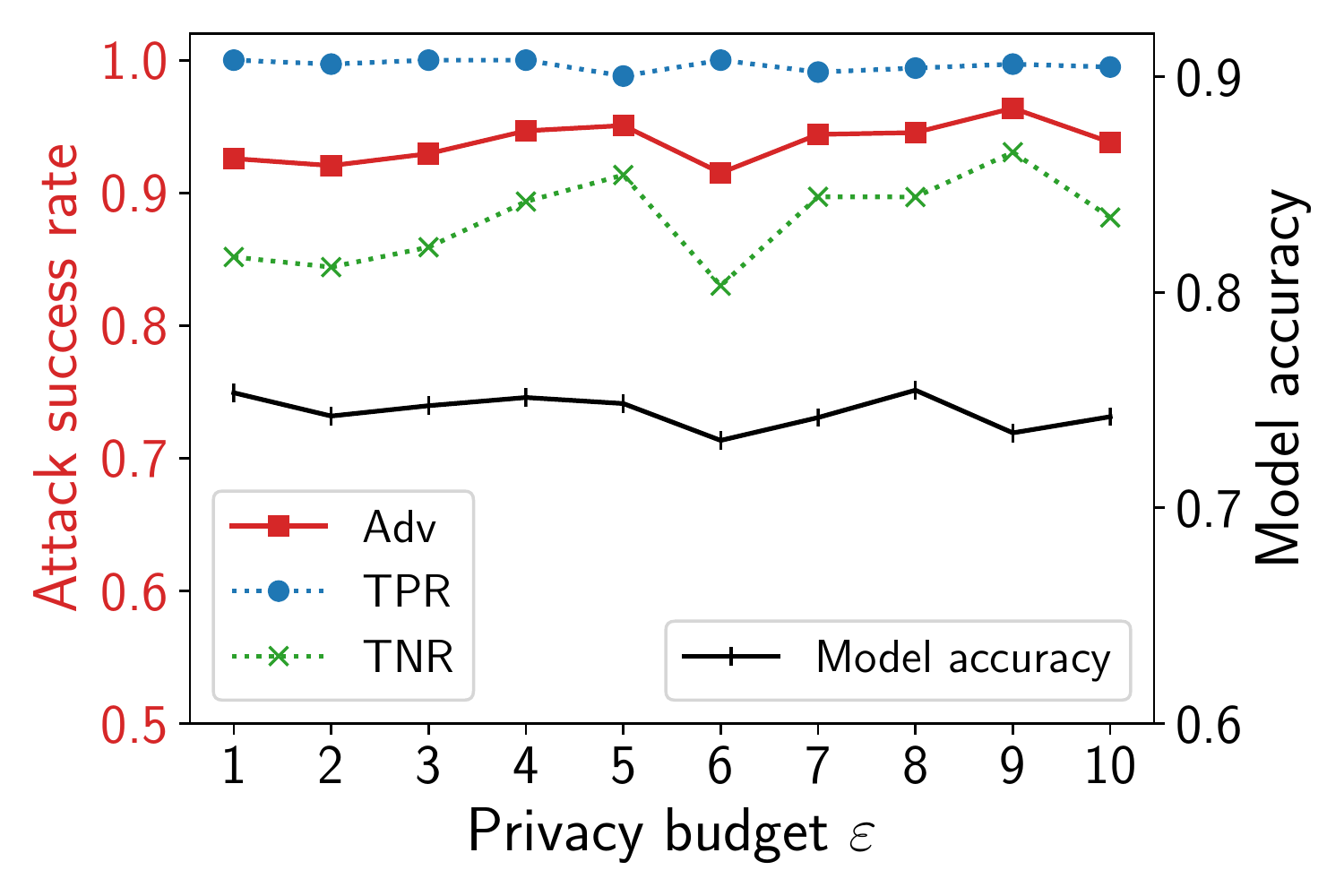}
        \caption{CIFAR-10}
    \end{subfigure}
    \caption{Attack success rate of AMI under the OME mechanism on CelebA, ImageNet, and CIFAR-10 datasets. The success rate is represented via the advantage (Adv), true positive rate (TPR), and true negative rate (TNR) according to Eq. \ref{equ:suc}. The baseline of random guessing is 0.5. The model accuracy illustrates the utility loss of the data when using LDP.}
    \label{fig:ome}
\end{figure*}

The model accuracies in Figs. \ref{fig:sucrate} and \ref{fig:ome} are measured by evaluating the model on legitimate classification tasks. For the CelebA dataset \cite{liu2015faceattributes}, the task is to classify whether a person is smiling or not based on face images. With regard to the CIFAR-10 dataset \cite{krizhevsky2009learning}, we use its original classification task with 10 classes. For the ImageNet dataset \cite{deng2009imagenet}, we extract a subset of 10 classes: tench, English springer, cassette player, chain saw, church, French horn, garbage truck, gas pump, golf ball, and parachute\footnote{\url{https://github.com/fastai/imagenette}}. Then, we evaluate the model performance on classifying those 10 classes. 

To obtain the feature embeddings of data samples, we use the pre-trained Resnet-18 model from Img2Vec\footnote{\url{https://github.com/christiansafka/img2vec}}.

\begin{figure*}[h]
    \centering
    \begin{subfigure}[b]{0.32\textwidth}
        \centering
        \includegraphics[width=\textwidth]{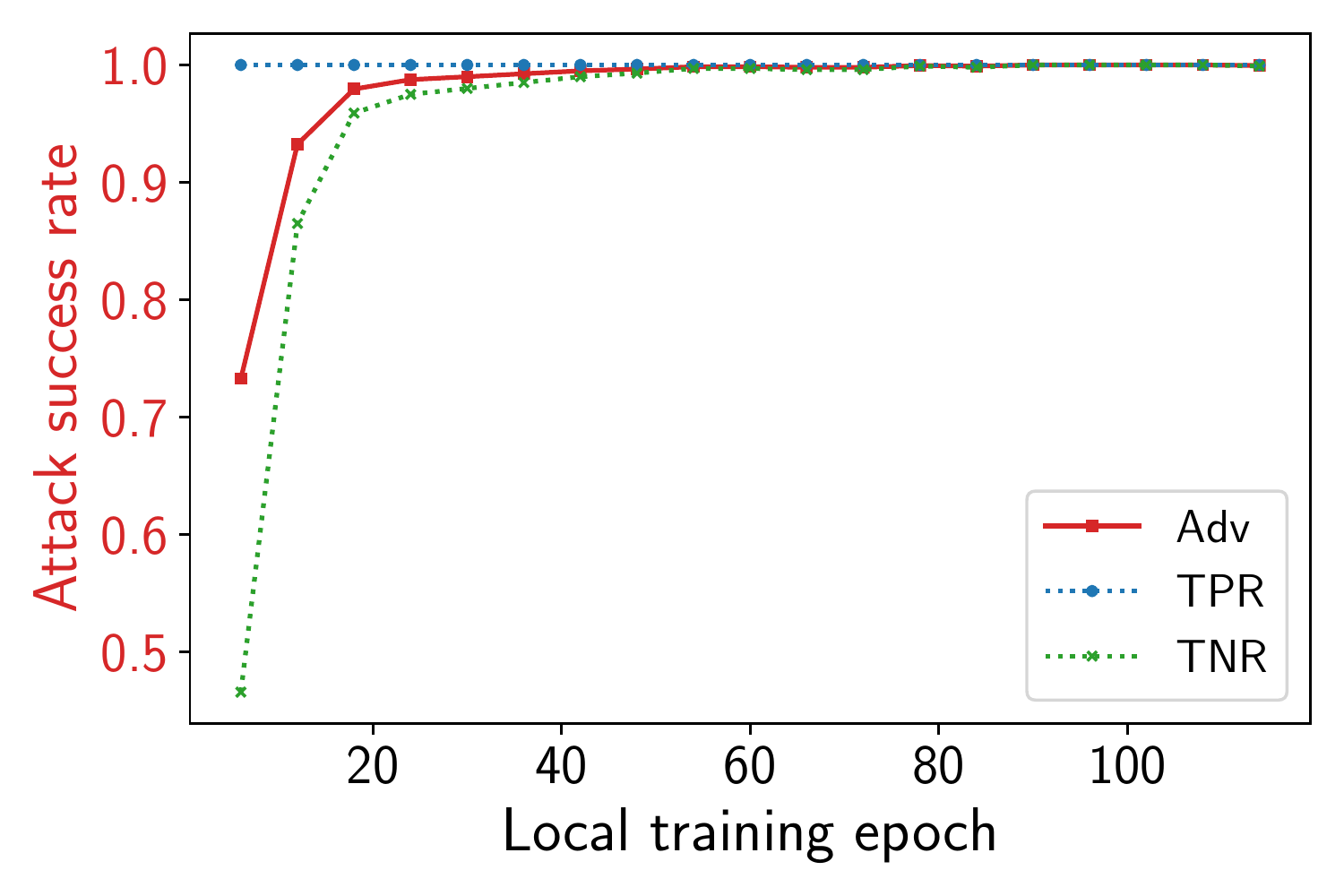}
        \caption{CelebA}
    \end{subfigure}
    \begin{subfigure}[b]{0.32\textwidth}
        \centering
        \includegraphics[width=\textwidth]{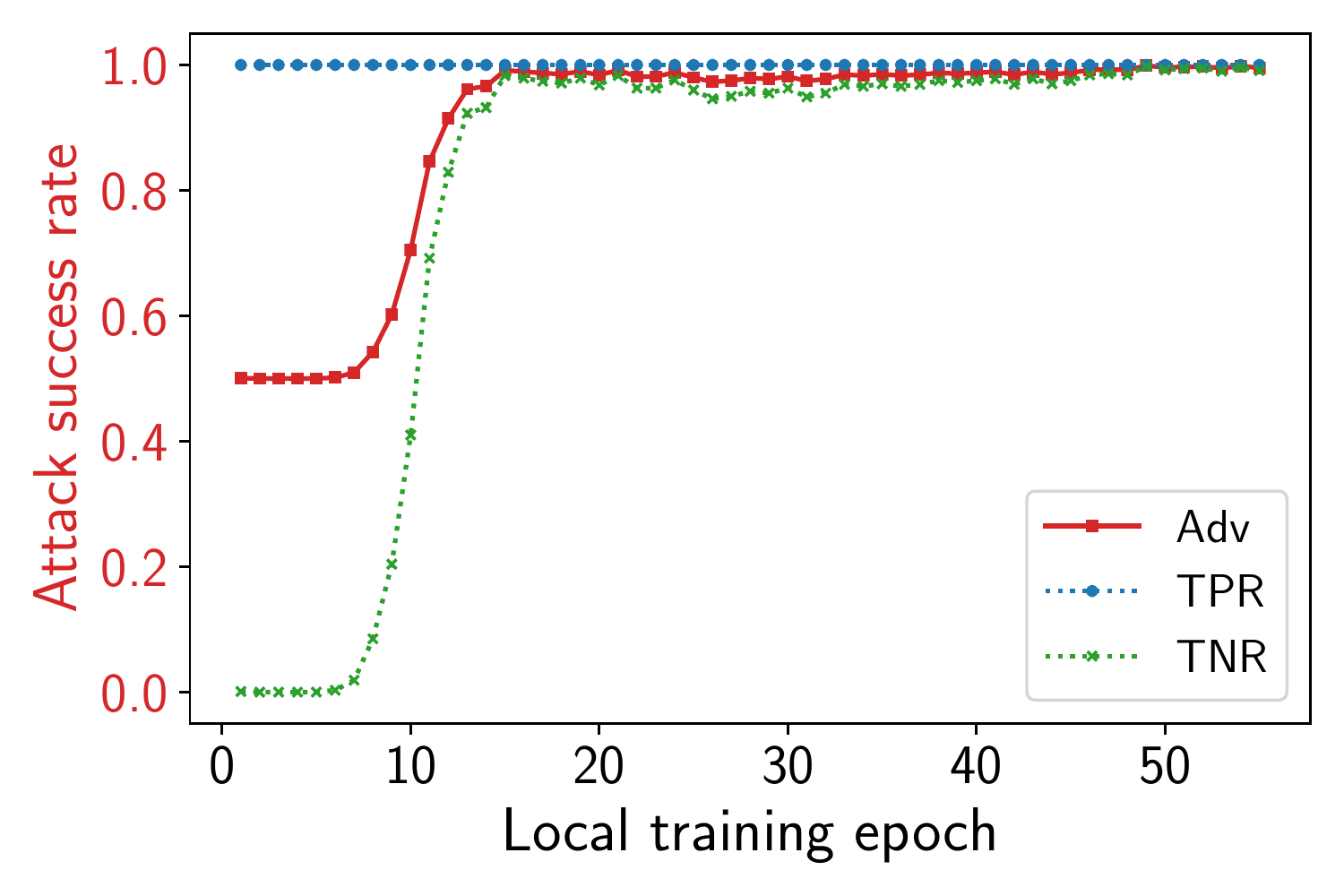}
        \caption{ImageNet}
    \end{subfigure}
    \begin{subfigure}[b]{0.32\textwidth}
        \centering
        \includegraphics[width=\textwidth]{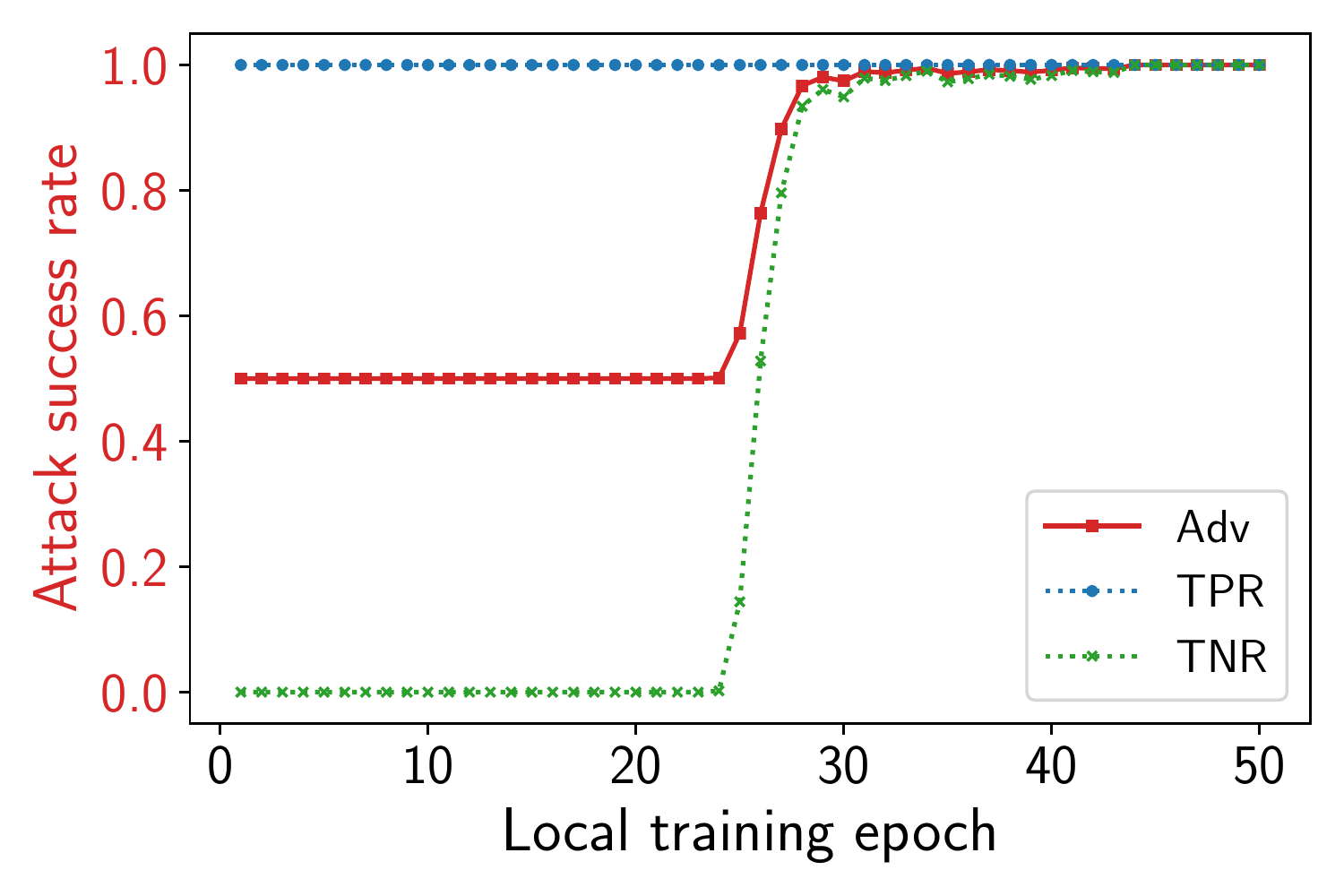}
        \caption{CIFAR-10}
    \end{subfigure}
    \caption{Attack success rate of AMI during local epochs of training $h, W$. The success rate is represented via the advantage (Adv), true positive rate (TPR), and true negative rate (TNR) according to Eq. \ref{equ:suc}. The baseline of random guessing is 0.5.}
    \label{fig:noldp}
\end{figure*}

\paragraph{Attack performance without LDP.} 
In this setting, the attack strategy in Fig. \ref{fig:strategy} requires the attacker to train the malicious parameters $h, W$, which takes multiple local training epochs. Fig. \ref{fig:noldp} shows the attack success rate per local epoch with 2,000 neurons in the first layer ($r=2,000$). As can be seen, over time, the attack success rate reaches 100\% across all three datasets. Table \ref{tab:epoch} shows the average number of local training epochs needed to train those parameters to achieve the 100\% success rate, as we vary $r$. We observe that increasing $r$ helps the attacker find the optimal parameters $h, W$ faster. For CIFAR-10, with $r=2,000$ neurons in the first layer, the attacker can easily train $h, W$ within 50 epochs.

To understand the reason behind this behavior, we note that training $h,W$ in this attack strategy is equivalent to finding a non-linear decision boundary that overfits to the target sample $t$ (Eq. \ref{equ:cond-nonlinear}), thereby distinguishing the target sample $t$ from any other samples. Hence, increasing $r$ raises the chance of over-fitting which, in turn, shortens the time to find $h, W$. Furthermore, Table \ref{tab:epoch} indicates that as few as $r=5$ neurons are needed to attain the 100\% success rate, albeit the longer training time.

\begin{table}[h]
\centering
\caption{Number of local epochs needed to train $h,W$  by the adversary $\calA$ (Fig. \ref{fig:strategy}) to obtain a 100\% success rate. We vary the number of neurons in the first layer ($r$) and get the average number of local epochs over multiple runs.}
\label{tab:epoch}
\begin{tabular}{|l|l|c|}
\hline
\textbf{Dataset} & \textbf{$r$} & \textbf{\begin{tabular}[c]{@{}l@{}}No. of local epochs 
\end{tabular}} \\ \hline
\multirow{4}{*}{CelebA}   & 5     & 3585 \\
                          & 500   & 763  \\
                          & 1000  & 497  \\
                          & 2000  & 297  \\ \hline
\multirow{4}{*}{ImageNet} & 5     & 1610 \\
                          & 500   & 131  \\
                          & 1000  & 88   \\
                          & 2000 & 63   \\ \hline
\multirow{4}{*}{CIFAR-10} & 5     & 309  \\
                          & 500   & 91   \\
                          & 1000  & 54   \\
                          & 2000  & 44   \\ \hline
\end{tabular}
\end{table}

\begin{figure*}[h]
    \centering
    \begin{subfigure}[t]{0.24\textwidth}
        \centering
        \includegraphics[width=\textwidth]{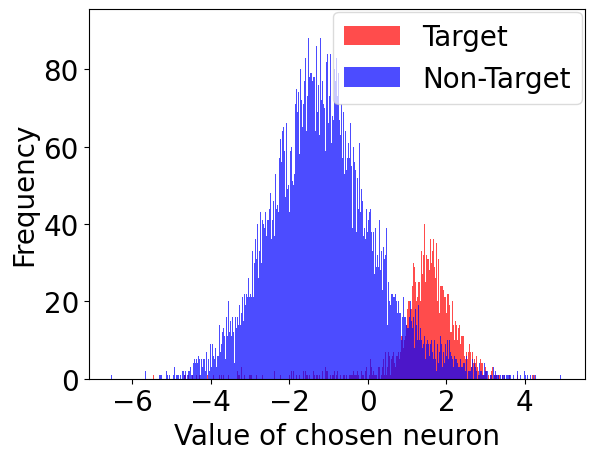}
        \caption{$\varepsilon=1$}
        \label{fig:hist-1}
    \end{subfigure}
    \begin{subfigure}[t]{0.24\textwidth}
        \centering
        \includegraphics[width=\textwidth]{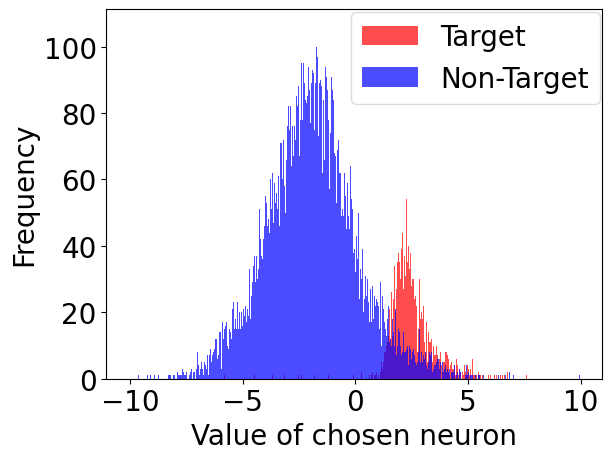}
        \caption{$\varepsilon=3$}
        \label{fig:hist-3}
    \end{subfigure}
    \begin{subfigure}[t]{0.24\textwidth}
        \centering
        \includegraphics[width=\textwidth]{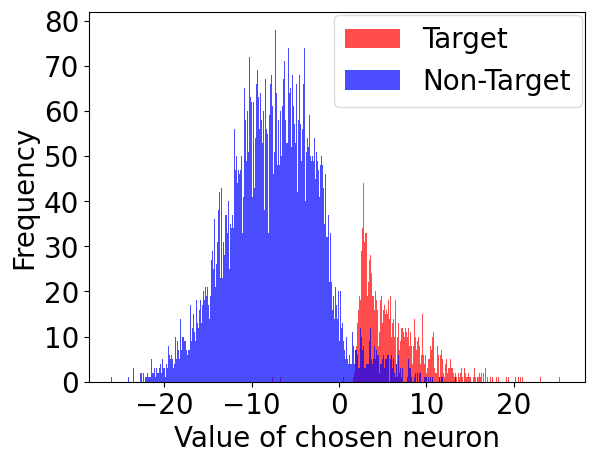}
        \caption{ $\varepsilon=5$}
        \label{fig:hist-5}
    \end{subfigure}
    \begin{subfigure}[t]{0.24\textwidth}
        \centering
        \includegraphics[width=\textwidth]{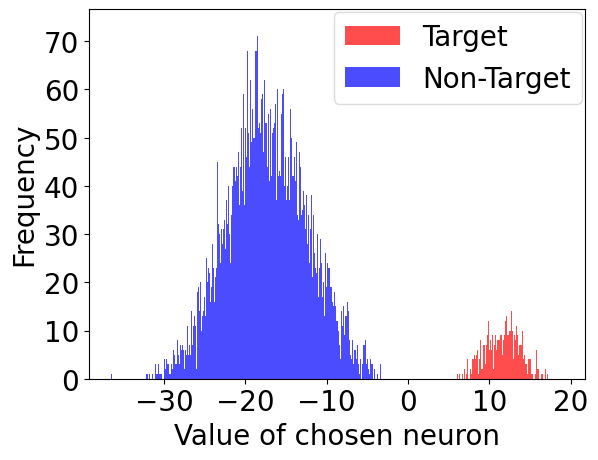}
        \caption{$\varepsilon=10$}
        \label{fig:hist-10}
    \end{subfigure}
    \caption{Histograms of the values of the chosen neuron  when attacking against the BitRand mechanism with $\varepsilon=1, 3, 5,$ and $10$, respectively. The higher $\varepsilon$, the less overlapping distribution of the values of the chosen neuron, given the  target samples and the non-target samples. This indicates the higher attack success rate of AMI. These data samples are obtained from the CelebA dataset. }
    \label{fig:histogram-neuron-value}
\end{figure*}

\paragraph{Attack performance under LDP.} To shed light into how the privacy budget $\varepsilon$ in the  LDP mechanism, i.e., BitRand,  affects the AMI success rate, we visualize the distribution of the values of the chosen neuron associated with the  target and non-target samples. In Fig.~\ref{fig:hist-1},  with rigorous privacy budget (e.g., $\varepsilon = 1$), the mean values of the chosen neuron is positive given the  target samples and  negative given the  non-target samples. However, there is a notable overlap in the two distributions of the  target samples (i.e., red distribution) and the non-target samples (i.e., blue distribution). This makes the attack difficult in distinguishing the target and non-target samples, based on the value of the chosen neuron. Consequently, the attack success rate is moderate. On the other hand, when $\varepsilon$ increases (Figs.~\ref{fig:hist-3}-\ref{fig:hist-10}), the distribution of the target samples  shift to the right, meanwhile the distribution of the non-target samples shift to the left. The shifts result in a less overlap between the two distributions. As a result, when the privacy budget $\varepsilon$ increases, the attack success rate of AMI increases. We observe this phenomenon in all three datasets. 

\begin{figure*}[t]
  \centering
\subfloat[CelebA]{\includegraphics[scale=0.28]{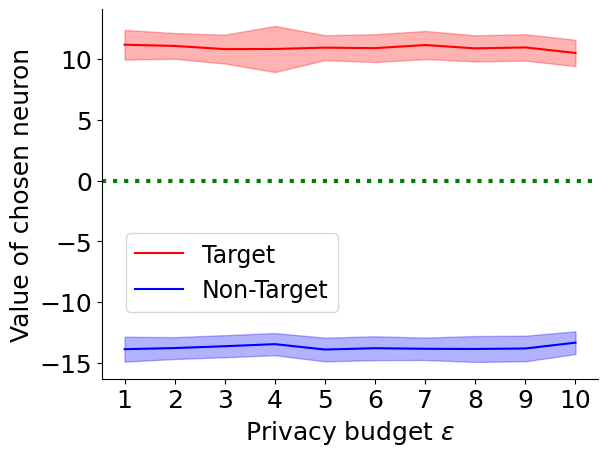}} 
\hspace{1cm}
\subfloat[ImageNet]{\includegraphics[scale=0.28]{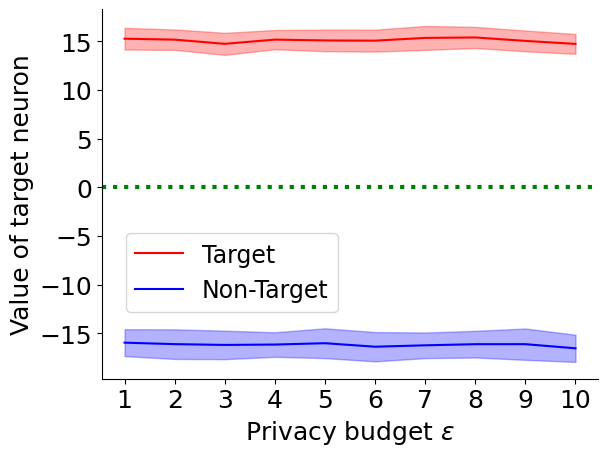}}\hspace{1cm}
\subfloat[CIFAR-10]{\includegraphics[scale=0.28]{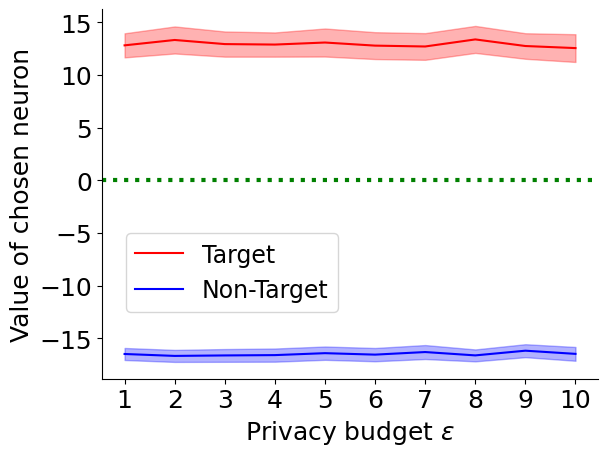}} 
      \caption{Certified guarantee of success for $\varepsilon \in [1,10]$ in OME \cite{lyu2020towards}: Expectation (solid lines), upper-bound, and lower-bound
(shaded areas surrounding the expectation) of the values of the chosen neuron. }
    \label{fig:boundOME}
 \end{figure*}

In addition to evaluating the attack under the BitRand mechanism in Section \ref{sec:eval}, Fig. \ref{fig:ome} shows our attack performance under the OME mechanism \cite{lyu2020towards}. First, we observe the same phenomenon of OME as in \cite{arachchige2019local, lyu2020towards} in which the model accuracy does not change much for $\varepsilon \in [1, 10]$. Second, our attack maintains a severe success rate of about 90\% for CelebA and CIFAR-10. More importantly, the attack success rate reaches more than 95\% for ImageNet. These results demonstrate that our AMI attack remains very effective even with low privacy budget $\varepsilon$ when the training set $\calD$ is protected by the OME mechanism.

Fig.~\ref{fig:boundOME} shows the certified guarantee of
success for the CelebA, ImageNet, and CIFAR-10 datasets when the LDP-preserving OME mechanism is used. We can derive a certified guarantee of success for our AMI attack given $\varepsilon \geq 1$ with a small broken probability $10^{-8}$. This result  is consistent with our attack success rate reported in Fig.~\ref{fig:ome}. In addition to the unaffected model accuracy and attack success rate, the effect of the privacy budget $\varepsilon$ is  modest in the certified guarantee of success, given $\varepsilon \in [1,10]$ used in OME.

\section{NOISY GRADIENTS WITH DPSGD} \label{sec:def}
Aside from LDP where clients perturb their own local training data $\calD$ before computing the gradients, another method is to let clients add DP noise to their gradients using DPSGD \cite{abadi2016deep} before sending them to the server, hindering the attacker from knowing the true value of the chosen neuron's gradient $g_t$. Nevertheless, recent work suggests that using DPSGD makes it impossible to train models with reasonable accuracy for datasets like CIFAR-10 or ImageNet, even in a non-distributed setting \cite{boenisch2021curious,tramer2020differentially}.



Moreover, even when DPSGD is used to add noise to the gradients, we can leverage the fact that the FL training is done in multiple iterations to circumvent this DP noise. In \cite{abadi2016deep}, the DP noise is sampled from a zero-mean Gaussian distribution with a standard deviation of $\sigma = \frac{\sqrt{2\ln(1.25/\delta)}}{\varepsilon}$, where $\varepsilon$ is the privacy budget and $\delta$ is a broken probability. However, as the noise is zero-mean, averaging the noise samples over multiple iterations will cancel out the noise and reveal the true value of the gradients. This is also referred to as the privacy composition problem in DP \cite{dwork2014algorithmic}.

\begin{figure}
    \centering
    \begin{subfigure}[b]{0.33\linewidth}
        \centering
        \includegraphics[width=\textwidth]{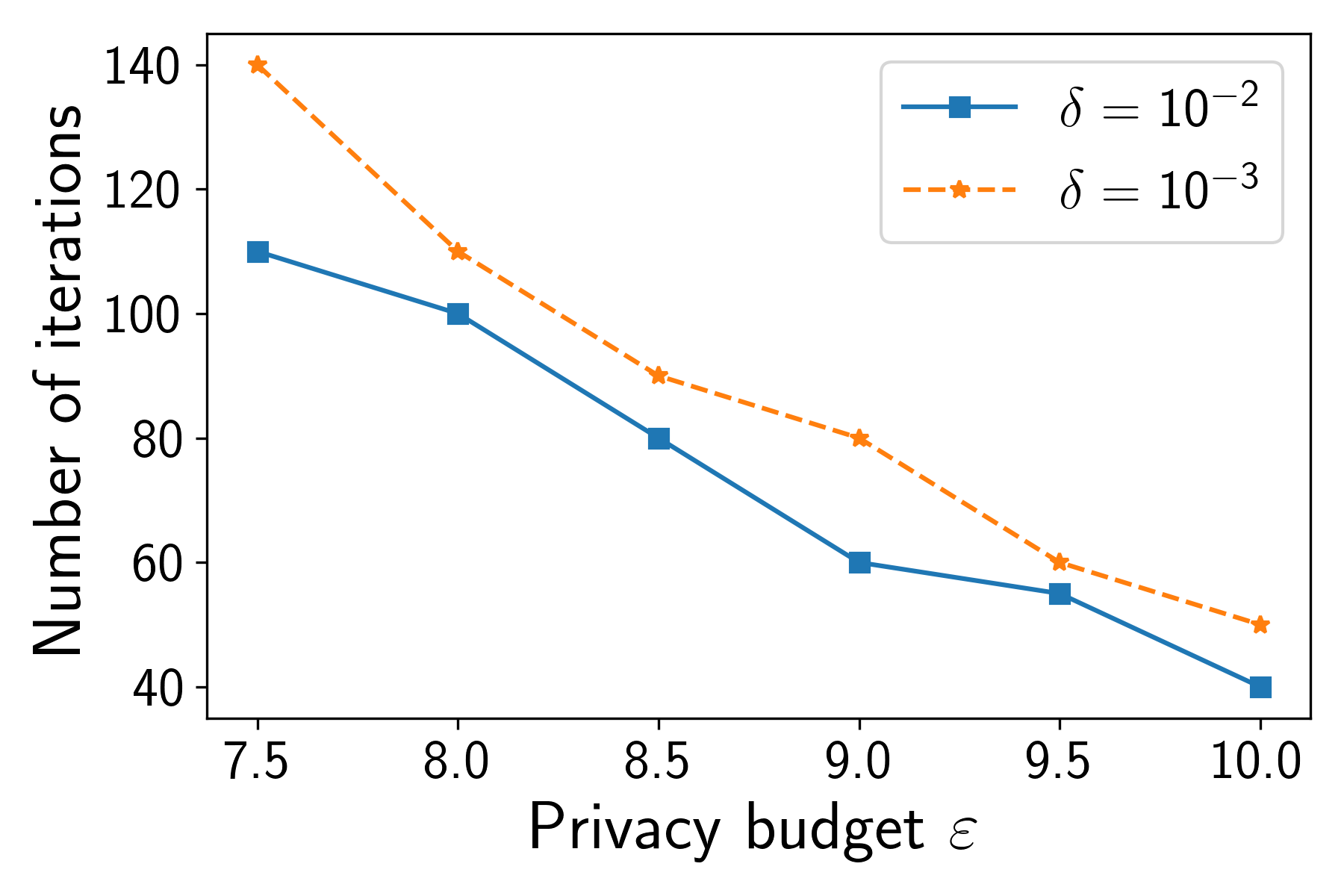}
        \caption{1 chosen neuron}
        \label{fig:dpsgd-1neuron}
    \end{subfigure}%
    \hspace{1cm}
    \begin{subfigure}[b]{0.33\linewidth}
        \centering
        \includegraphics[width=\textwidth]{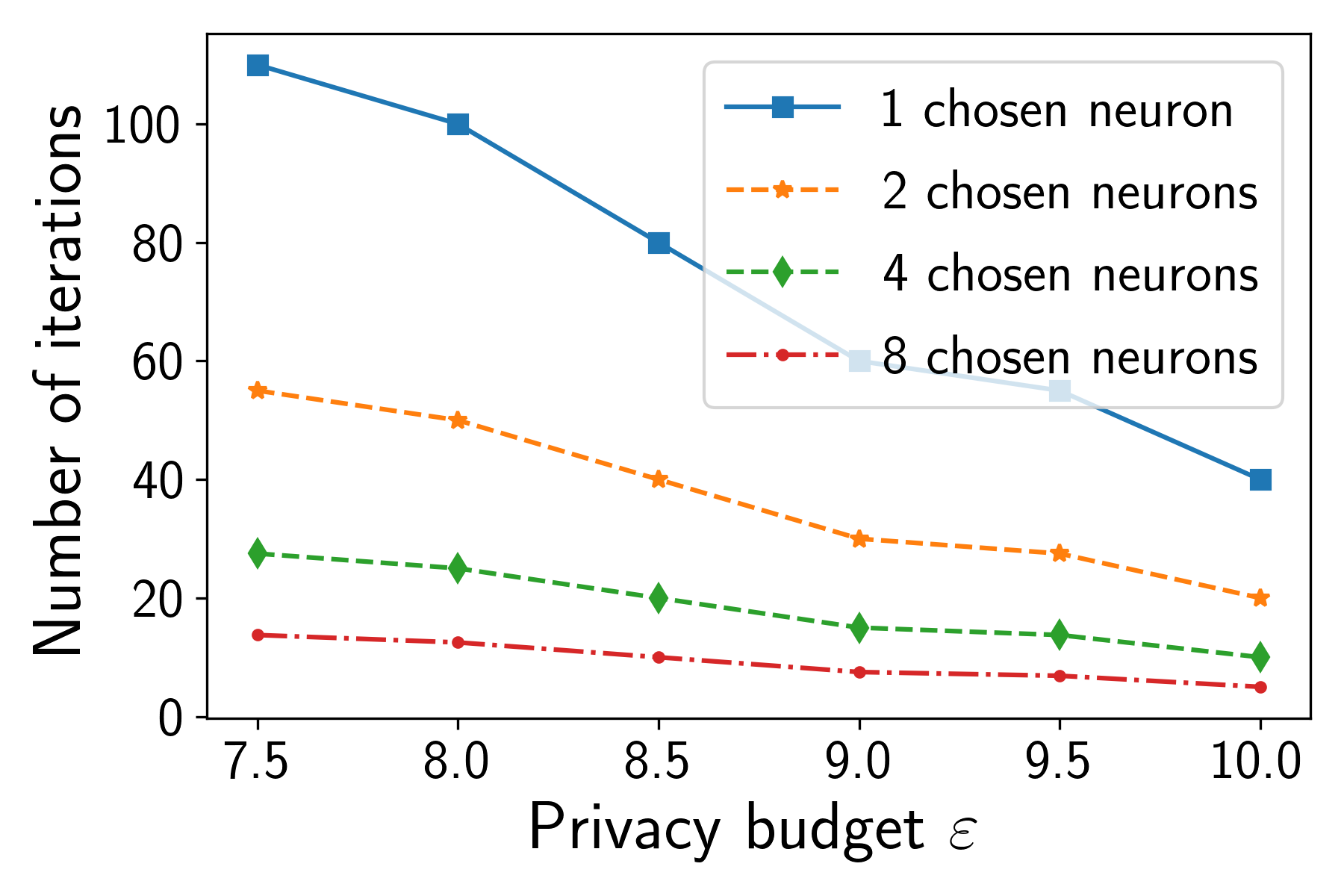}
        \caption{$\delta = 10^{-2}$}
        \label{fig:dpsgd-multineuron}
    \end{subfigure}
    \caption{Number of iterations needed to eliminate DP noises. \ref{fig:dpsgd-1neuron} shows the results for 1 chosen neuron, while \ref{fig:dpsgd-multineuron} varies the number of chosen neurons and fixes $\delta = 10^{-2}$.}
    \label{fig:dpnoise}
\end{figure}

Specifically, denoting $g_t^{(i)'} = g_t + z_i$ (where $z_i \sim \calN(0, \sigma^2 \mathbf{I})$) as the noisy gradient of the chosen neuron at iteration $i$, the server can obtain the true gradient $g_t$ by averaging over multiple iterations, i.e., $g_t = \frac{1}{P} \sum_{i=1}^P g_t^{(i)'}$. From this $g_t$, the attacker can determine whether the target sample was used in at least one of those iterations, following the same principle in Section \ref{sec:amia}. Fig. \ref{fig:dpsgd-1neuron} shows the number of iterations needed to eliminate the DP noise. Previous work shows that training a neural network for CIFAR-10 up to a modest accuracy of 66.2\% requires a privacy budget $\varepsilon \geq 7.53$ \cite{tramer2020differentially}, hence we evaluate with $\varepsilon \geq 7.5$. When $\varepsilon \geq 8$, we need less than 100 iterations.

To reduce the number of iterations, we can increase the number of chosen neurons in the second layer, and average the noisy gradients over all of the chosen neurons. Simply speaking, having $K$ chosen neurons would reduce the number of iterations by $K$-fold. Fig. \ref{fig:dpsgd-multineuron} shows the number of iterations needed to eliminate the DP noise with multiple neurons. As can be seen, with only 4 chosen neurons, we only need less than 60 iterations to cancel out the DP noise at $\varepsilon=7.5$. With 8 chosen neurons, the noise can be canceled out within 10 iterations.

\begin{figure*}[]
    \centering
    \begin{subfigure}[t]{0.24\textwidth}
        \centering
        \includegraphics[width=\textwidth]{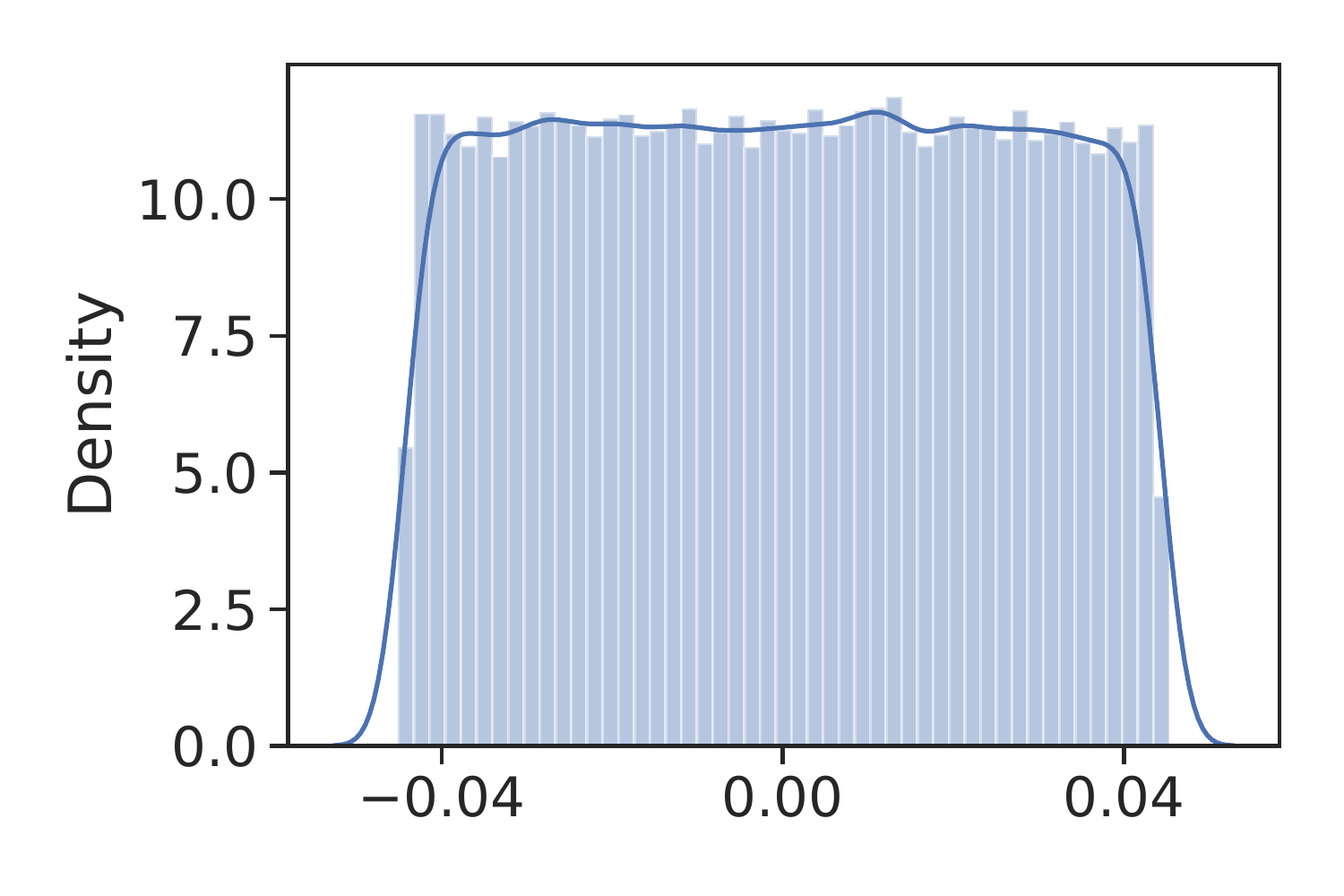}
        \caption{Normal weights}
        \label{fig:normal}
    \end{subfigure}
    \begin{subfigure}[t]{0.24\textwidth}
        \centering
        \includegraphics[width=\textwidth]{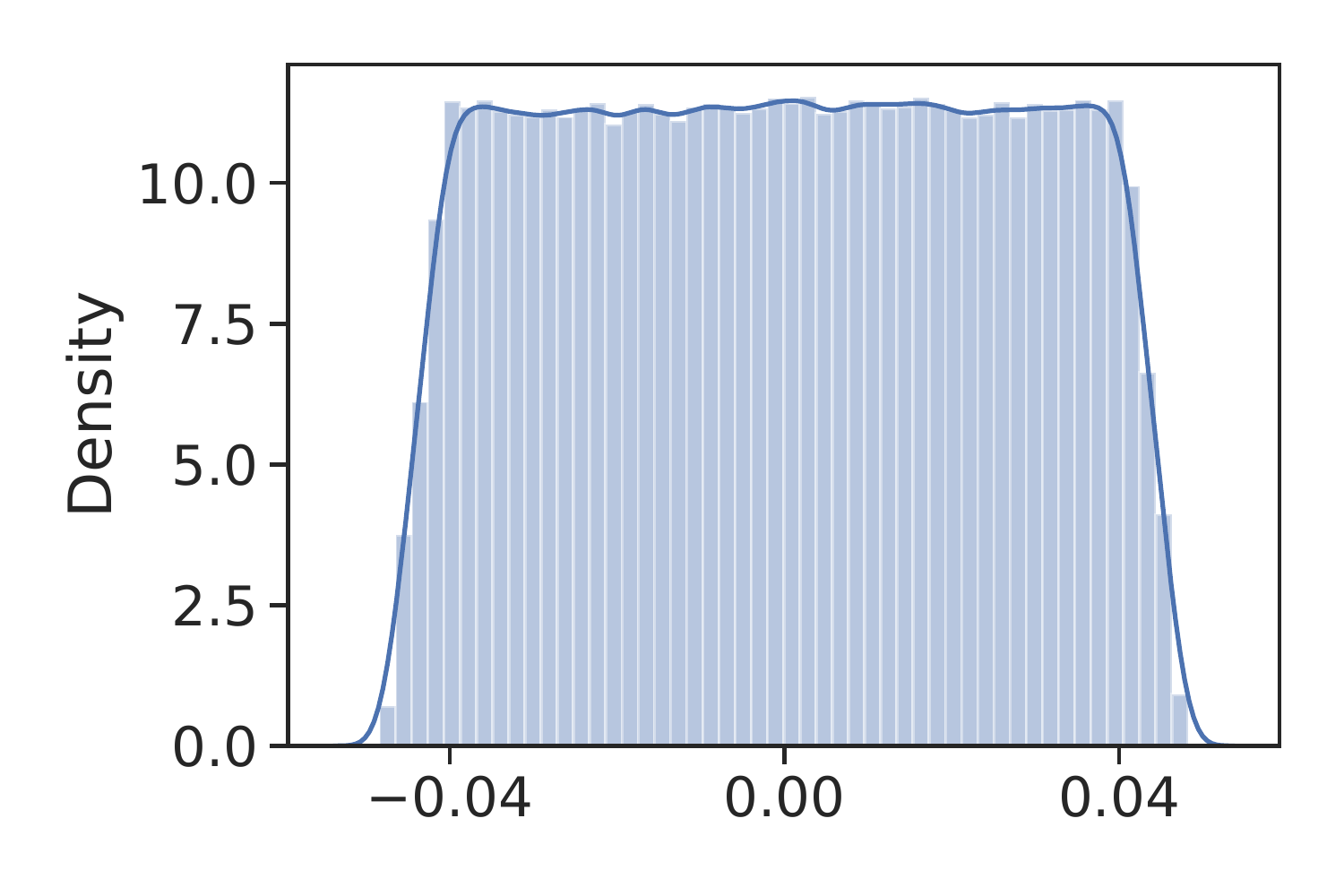}
        \caption{$\varepsilon=2$}
        \label{fig:mal2}
    \end{subfigure}
    \begin{subfigure}[t]{0.24\textwidth}
        \centering
        \includegraphics[width=\textwidth]{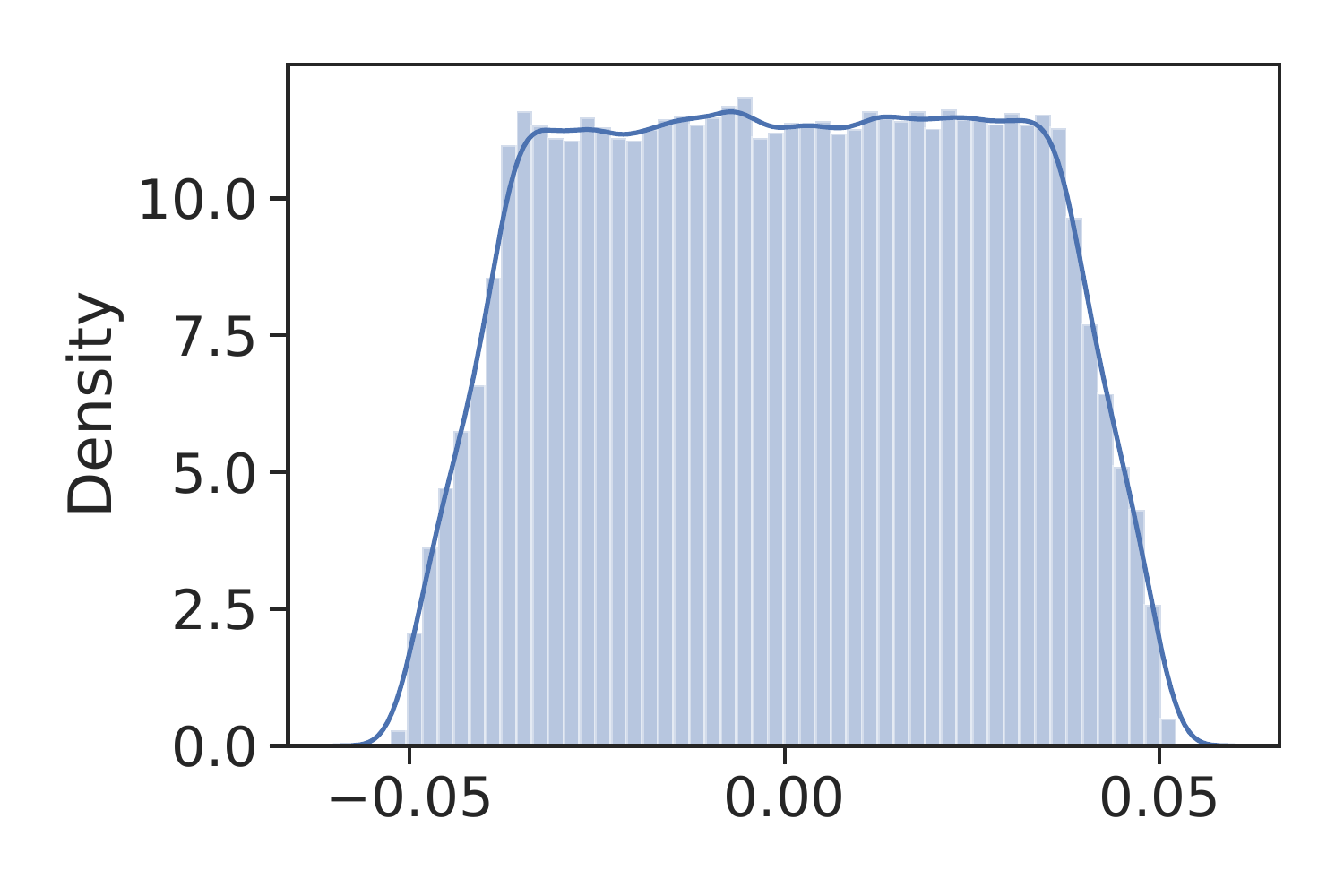}
        \caption{ $\varepsilon=3$}
        \label{fig:mal3}
    \end{subfigure}
    \begin{subfigure}[t]{0.24\textwidth}
        \centering
        \includegraphics[width=\textwidth]{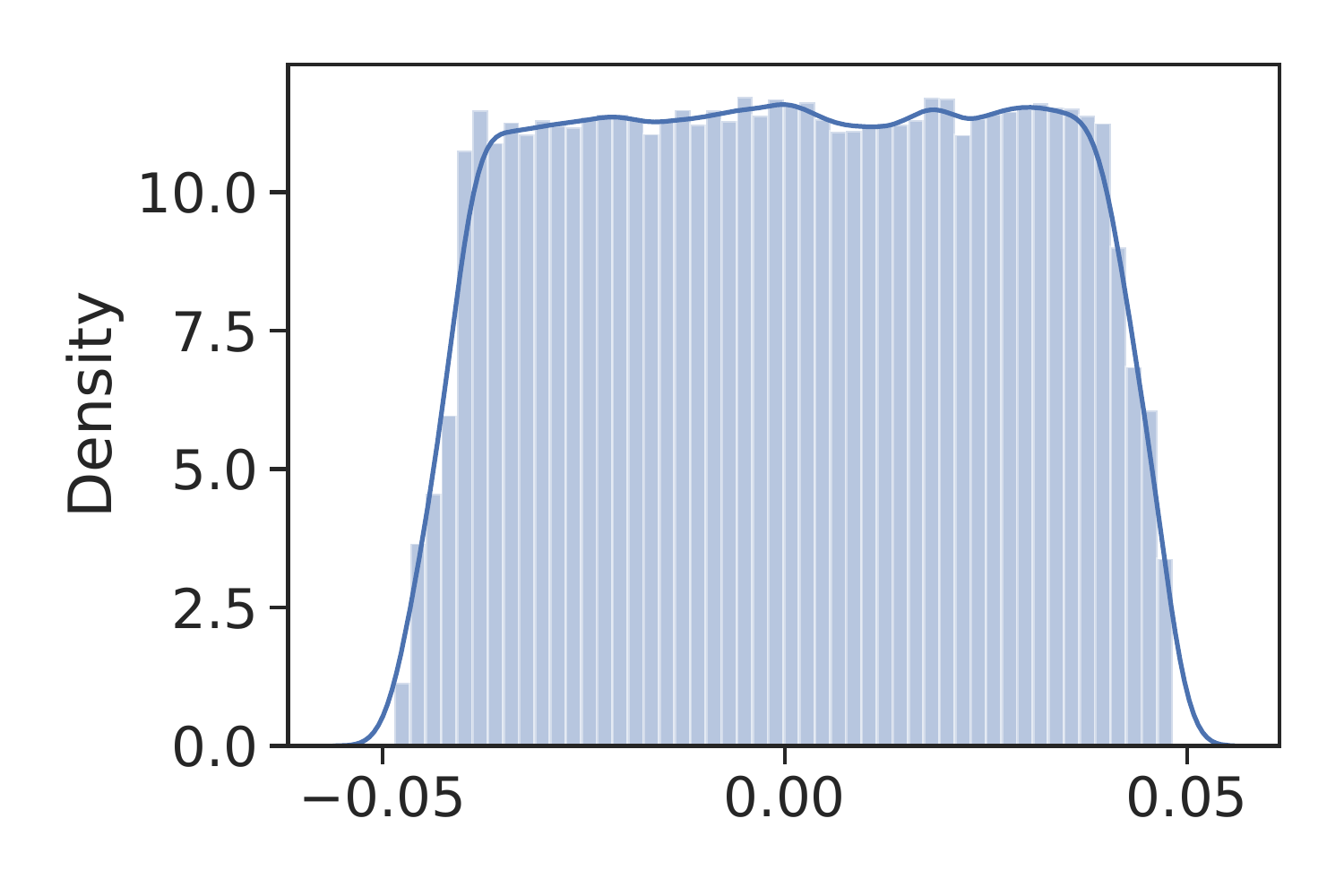}
        \caption{$\varepsilon=5$}
        \label{fig:mal5}
    \end{subfigure}
    \caption{Histograms of the distribution of normal weights and malicious weights. \ref{fig:normal} shows the normal weights, while \ref{fig:mal2}, \ref{fig:mal3}, and \ref{fig:mal5} show the malicious weights when attacking under BitRand with $\varepsilon=2,3,$ and $5$, respectively. The solid blue line is the kernel density estimation (KDE).}
    \label{fig:dist}
\end{figure*}

\end{document}